\documentclass[10pt,journal,compsoc]{IEEEtran}

\ifCLASSOPTIONcompsoc
  \usepackage[nocompress]{cite}
\else
  \usepackage{cite}
\fi

\ifCLASSINFOpdf

\else

\fi

\usepackage{epsfig}
\usepackage{graphicx}
\usepackage{amsmath}
\usepackage{amssymb}
\usepackage{booktabs}
\usepackage{comment}
\usepackage{multirow}

\usepackage{booktabs}
\usepackage{array, caption, threeparttable}
\usepackage{caption}

\usepackage{algorithm}
\usepackage{listings}
\usepackage{color}

\usepackage{mathtools}

\usepackage{todonotes}
\usepackage{gensymb}

\usepackage{hyperref}
\hypersetup{
    colorlinks=true,
    linkcolor=blue,
    filecolor=magenta,      
    urlcolor=cyan,
}

\begin{document}

\title{NAS-FAS: Static-Dynamic Central Difference Network Search for Face Anti-Spoofing}

\author{Zitong Yu,~\IEEEmembership{Student Member,~IEEE}, Jun Wan,~\IEEEmembership{Senior Member,~IEEE}, Yunxiao Qin, \\ Xiaobai Li,~\IEEEmembership{Member,~IEEE}, 
Stan Z. Li,~\IEEEmembership{Fellow,~IEEE} and Guoying Zhao,~\IEEEmembership{Senior Member,~IEEE}


\IEEEcompsocitemizethanks{\IEEEcompsocthanksitem Z. Yu, X. Li and G. Zhao are with Center for Machine Vision and Signal Analysis, University of Oulu, Oulu 90014, Finland.

E-mail: \{zitong.yu, xiaobai.li, guoying.zhao\}@oulu.fi.

\IEEEcompsocthanksitem J. Wan is with the National Laboratory of Pattern Recognition, Institute of Automation, Chinese Academy of Sciences, and University of Chinese Academy of Sciences, Beijing 100190, China. E-mail: jun.wan@ia.ac.cn

\IEEEcompsocthanksitem Y. Qin is with Northwestern Polytechnical University, Xian 710072, China.
E-mail: qyxqyx@mail.nwpu.edu.cn.

\IEEEcompsocthanksitem Stan Z. Li is with School of Engineering, Westlake University, Hangzhou 310012, China. E-mail: stan.zq.li@westlake.edu.cn.

}
\thanks{Manuscript received March 18, 2020; revised August 9 and September 25, 2020; accepted October 26, 2020 (Corresponding authors: Guoying Zhao and Jun Wan)}}

\markboth{IEEE TRANSACTIONS ON PATTERN ANALYSIS AND MACHINE INTELLIGENCE}%
{Shell \MakeLowercase{\textit{et al.}}: Bare Advanced Demo of IEEEtran.cls for IEEE Computer Society Journals}

\IEEEtitleabstractindextext{%
\begin{abstract}

Face anti-spoofing (FAS) plays a vital role in securing face recognition systems. Existing methods heavily rely on the expert-designed networks, which may lead to a sub-optimal solution for FAS task. Here we propose the first FAS method based on neural architecture search (NAS), called NAS-FAS, to discover the well-suited task-aware networks. Unlike previous NAS works mainly focus on developing efficient search strategies in generic object classification, we pay more attention to study the search spaces for FAS task. The challenges of utilizing NAS for FAS are in two folds: the networks searched on 1) a specific acquisition condition might perform poorly in unseen conditions, and 2) particular spoofing attacks might generalize badly for unseen attacks. To overcome these two issues, we develop a novel search space consisting of central difference convolution and pooling operators. Moreover, an efficient static-dynamic representation is exploited for fully mining the FAS-aware spatio-temporal discrepancy. Besides, we propose Domain/Type-aware Meta-NAS, which leverages cross-domain/type knowledge for robust searching. Finally, in order to evaluate the NAS transferability for cross datasets and unknown attack types, we release a large-scale 3D mask dataset, namely CASIA-SURF 3DMask, for supporting the new ‘cross-dataset cross-type’ testing protocol. Experiments demonstrate that the proposed NAS-FAS achieves state-of-the-art performance on nine FAS benchmark datasets with four testing protocols.



\end{abstract}

}

\maketitle

\IEEEdisplaynontitleabstractindextext

%
\IEEEpeerreviewmaketitle

\ifCLASSOPTIONcompsoc
\IEEEraisesectionheading{\section{Introduction}\label{sec:introduction}}
\else
\section{Introduction}
\label{sec:introduction}
\fi

\IEEEPARstart{F}{ace} recognition technology has become the most indispensable component in many interactive intelligent systems due to their convenience and remarkable accuracy. However, most existing face recognition systems are vulnerable to presentation attacks (PAs) ranging from print, replay and 3D-mask attacks. Therefore, not only the academia but also the industry has recognized the critical role of face anti-spoofing (FAS) for securing the face recognition system.

\begin{figure}
\centering
\includegraphics[scale=0.51]{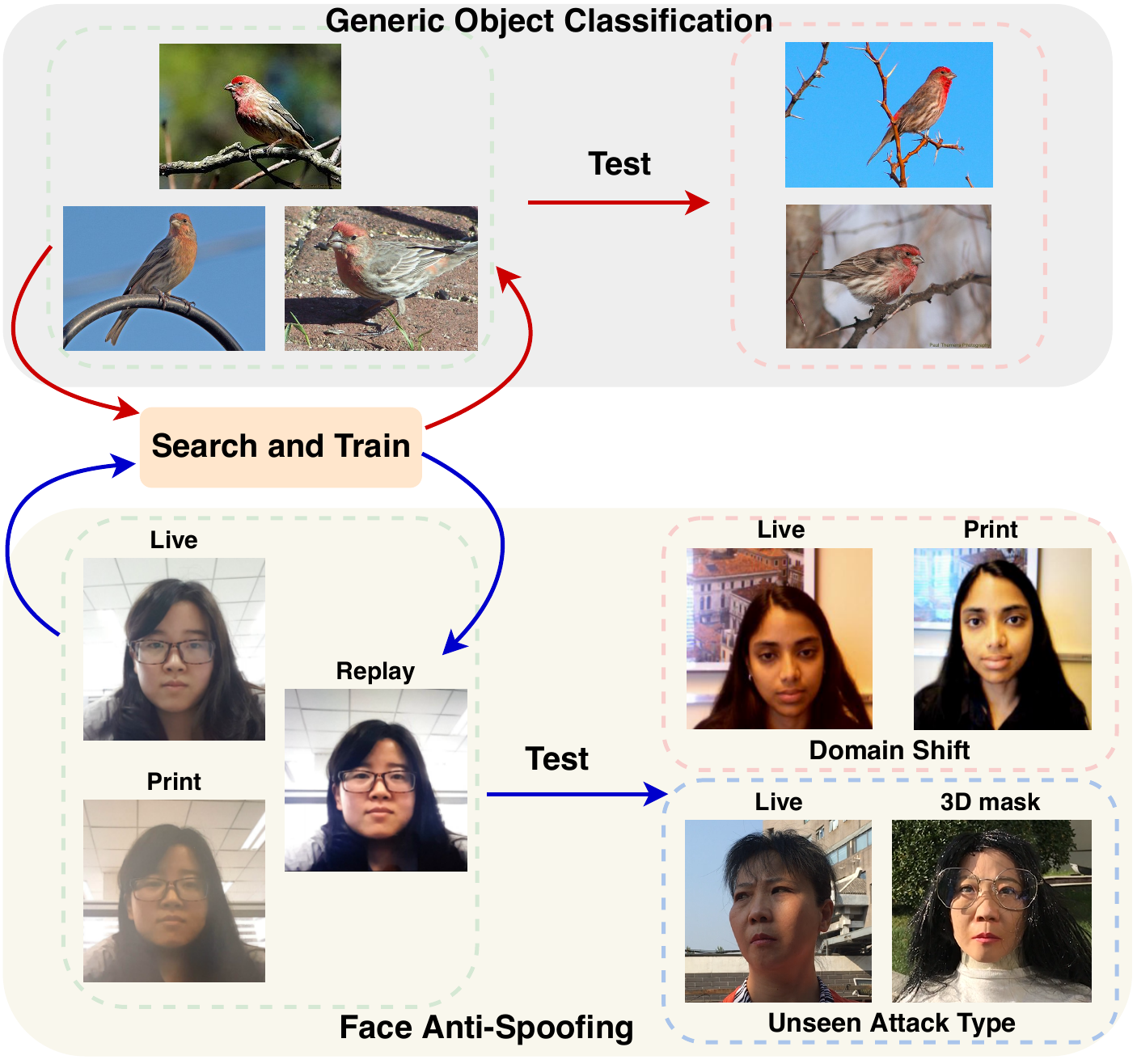}
\vspace{-0.6em}
  \caption{ 
  Compared with generic object classification task, the challenge of using neural architecture search for the FAS task derives from domain shift and unseen spoof attacks. 
  }
\label{fig:Figure1}
\vspace{-0.6em}
\end{figure}

\begin{figure}
\centering
\includegraphics[scale=0.14]{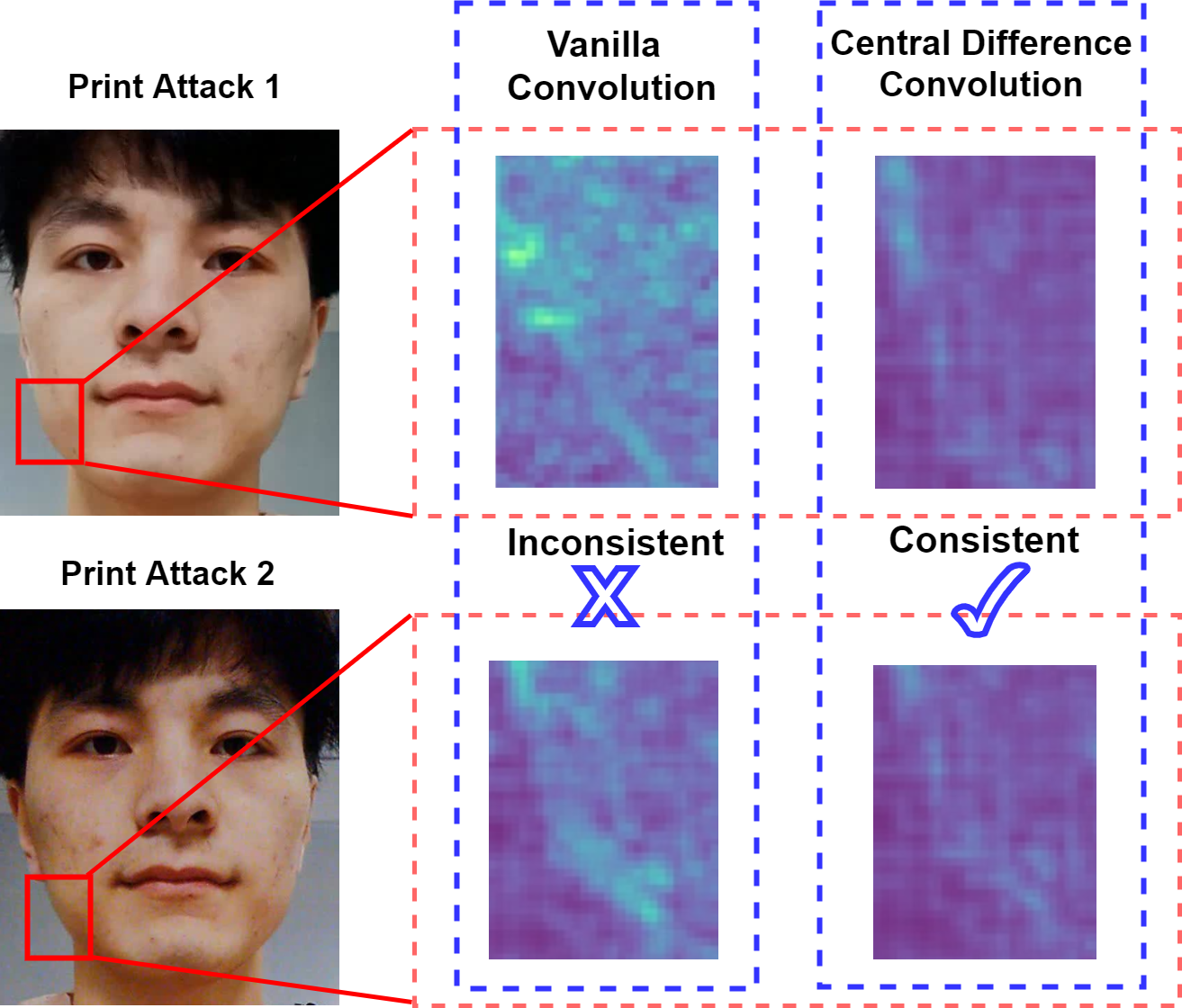}
  \caption{
  Feature responses of vanilla convolution (VanillaConv) and CDC for spoofing faces in shifted domains (illumination \& input camera). Consistent spoofing pattern (e.g., lattice artifacts) is observed using the CDC.
  }
\label{fig:Figure2}
\vspace{-0.6em}
\end{figure}

In the past few years, both traditional~\cite{Pereira2012LBP,Komulainen2014Context,Patel2016Secure} and deep learning-based~\cite{yu2020searching,yu2020face,Liu2018Learning,jourabloo2018face,yang2019face,Atoum2018Face,yu2020multi} methods have shown effectiveness for presentation attack detection (PAD). On one hand, some classical local descriptors (e.g., local binary pattern (LBP)~\cite{boulkenafet2015face} and histogram of gradient (HOG)~\cite{Komulainen2014Context}) are robust for describing the detailed invariant information (e.g., color texture, moir$\rm\acute{e}$ pattern and noise artifacts) from spoofing faces. However, the shallow and coarse feature extraction procedure limits the discriminative capacity of these local descriptors. On the other hand, convolutional neural networks (CNNs) focus on representing deeper semantic features to distinguish the bona fide and PA, which are weak in capturing fine-grained intrinsic patterns (e.g., lattice artifacts shown in Fig.~\ref{fig:Figure2}) between live and spoofing faces, and easily influenced by the variant scenarios. In consideration of the representational advantages of the local descriptors (detailed and robust) and CNNs (semantic and discriminative), \textbf{it is worth exploring the integration between local descriptors with convolution/pooling operators for robust and discriminative FAS.}


Although static spatial information plays key roles in FAS task, temporal/dynamic clue also contributes to robust feature representation, which could be revealed from the discrepancy (e.g., dynamic texture~\cite{tirunagari2015detection}, temporal depth~\cite{wang2018exploiting} and motion blurriness~\cite{li2019replayed}) between live and spoofing faces. However, existing methods usually adopt 3D convolution or long short-term memory modules for computing dynamic features, which needs extra network costs but with poor visual interpretation. \textbf{Designing a compact static-dynamic representation with visual interpretation (see Fig.~\ref{fig:STmap} for visual evidence) would be helpful to understand and tackle the FAS task}.

The classical backbones (e.g., VGG~\cite{simonyan2014very}, ResNet~\cite{he2016deep} and DenseNet~\cite{huang2017densely}) are first designed for generic object classification task, and transferred to the FAS task~\cite{yang2014learn,yang2019face,george2019deep}. However, all these backbones are carefully designed by human experts and lack of FAS task-oriented prior knowledge, which might not be optimal for FAS task. It is natural to think about the neural architecture search (NAS) with FAS-aware knowledge. For instance, with traditional cross-entropy loss, networks easily learn the arbitrary patterns such as screen bezel instead of the essential spoof patterns~\cite{Liu2018Learning}. In contrast, dense pixel-wise supervision signals such as pseudo depth map~\cite{Atoum2018Face,Liu2018Learning} or binary mask~\cite{george2019deep} are more helpful for learning detailed spoof cues. Hence, \textbf{valuable task-aware knowledge (e.g., supervision signals and search space design) should be considered in searching well-suited networks for FAS task.}


In generic object classification task, NAS is usually utilized to discover well-suited networks on a small proxy set (e.g., CIFAR-10) and then re-train and test on a large target set (e.g., ImageNet~\cite{deng2009imagenet}) (or search and test on target set directly). In other words, the data distribution in searching or re-training stage is similar to that in testing stage. However for FAS task, domain shift and unseen spoofing attack types occur universally. As a result, it is difficult to search and train a robust network in a source domain with limited seen attack types but test in a target domain with unseen attacks. The challenges are illustrated in Fig.~\ref{fig:Figure1}. In this paper, domain shift can be defined as two cases: 1) slight intra-dataset domain shift (e.g., changes of illumination, camera and face pose/expression), and 2) serious domain shift derived from cross datasets. It is interesting to study \textbf{how can the NAS methods be applied to search robust networks against domain shift and unseen attack types in FAS task?} 

Motivated by the discussions above, we propose the novel convolution and pooling operators called Central Difference Convolution (CDC) and Central Difference Pooling (CDP), which are good at describing fine-grained invariant information. As shown in Fig.~\ref{fig:Figure2}, CDC is more suitable to extract intrinsic spoofing patterns (e.g., lattice artifacts) than vanilla convolution in diverse environments. Furthermore, compact static-dynamic representation is developed for providing rich spatio-temporal discrepancy clues between live and spoofing faces. Finally, over a specifically designed task-aware search space, NAS is utilized to discover the robust static-dynamic networks against domain shift and unseen attacks in FAS task. 

This paper is an extended version of our prior publication~\cite{yu2020searching} in CVPR 2020. The main differences with the conference version are as follows: 1) besides the CDC, we propose CDP to form a unified CD-based family (both convolution and pooling operators) for FAS task; 2) unlike~\cite{yu2020searching} treating FAS as a static problem, we explore dynamic cues and design more powerful but efficient static-dynamic representation; 3) one more classical search space (i.e., baseline search space) is compared and discussed; 4) Domain/Type-aware Meta-NAS is proposed for efficiently searching on multiple domains/types; and 5) a new large-scale 3D mask dataset 'CASIA-SURF 3DMASK', and 'cross-dataset cross-type' testing protocols are established. To sum up, the main contributions of this paper are listed:

\begin{itemize}
\setlength\itemsep{-0.1em}
    
    
    \item We propose NAS-FAS, the first NAS approach for FAS task, to tackle the problems of domain shift and unseen attacks. Meanwhile, sufficient analyses about fine-grained NAS components (i.e., search space and supervision signals) in both static and static-dynamic FAS views are explored.
    \item  The central difference family (including CDC and CDP operators) is proposed for representing more intrinsic and robust FAS features. It proves that without CDC or CDP, the searched network performs poorly for the FAS task.
   \item  We propose the Domain/Type-aware Meta-NAS which is able to search generalized architectures via efficiently exploiting the domain/type shifted knowledge. To our best knowledge, this is the first work to search on multiple datasets.
    \item The proposed ‘cross-dataset cross-type’ testing protocol is first studied in FAS, which is used for evaluating the NAS transfer and generalization ability for both unseen domains and spoofing attack types. Furthermore, we release a large-scale 3D mask dataset, namely CASIA-SURF 3DMask, which is built up for supporting this challenging protocol.
    
    
    \item The proposed method has been evaluated on nine FAS benchmark datasets with four testing protocols, and achieves the state-of-the-art performance.
\end{itemize}

In the rest of the paper, Section~\ref{sec:relatedwork} provides the related work and Section~\ref{sec:method} formulates the central difference convolution and pooling operations and then describes the compact static-dynamic representation. Section~\ref{sec:nas} introduces the NAS methods with task-aware search space. Section~\ref{sec:dataset} gives details about our released CASIA-SURF 3DMASK as well as the existing datasets, and introduces four testing protocols. Section~\ref{sec:experiment} provides rigorous ablation studies and evaluates the performance of the proposed models on nine benchmark datasets. Finally, a conclusion is given in Section~\ref{sec:conclusion}.

\section{Related Work} \label{sec:relatedwork}
\noindent\textbf{Neural Architecture Search.}\quad   
As designing a high performance neural architecture requires substantial efforts and expertise, NAS becomes more and more important to discover best-suited networks automatically. The existing NAS methods could be summarized as these three categories: 1) Reinforcement learning based methods~\cite{zoph2016neural,zoph2018learning}; 2) Evolution algorithm based methods~\cite{real2019regularized,real2017large,real2019regularized}; and 3) Gradient based methods~\cite{liu2018darts,chen2019progressive,xu2019pc,cai2019proxylessnas,yao2020efficient}. Recently, several NAS benchmarks for generic object classification task such as NAS-Bench-101\cite{ying2019bench}, NAS-Bench-201\cite{dong2020bench}, and NAS-Bench-1Shot1\cite{zela2020bench} as well as evaluation manner\cite{yang2019evaluation} are proposed for fair performance comparison. In other side, in order to quickly adapt and discover excellent architectures in the unseen scenarios, some meta NAS based methods\cite{lian2019towards,elsken2020meta,shaw2019meta,kim2018auto,wang2020m} are developed. However, the existing meta NAS methods usually 1) need few target tasks for fast adaptation; and 2) only consider searching on a single dataset (domain). Thus, they are not suitable to search architectures for the domain generalization/open-set FAS tasks where the unknown target scenarios/attack types are unaccessible.

For the perspective of automated computer vision (AutoCV) applications, NAS has been developed for face analysis \cite{yu2020autohr}, gesture recognition \cite{yu2020searching2}, person ReID \cite{quan2019auto} and object detection \cite{ghiasi2019fpn} tasks. Different from generic object classification task, the FAS task relies on intrinsic cues between live and spoofing faces, which are easily contaminated by domain shift and unknown attack types. To the best of our knowledge, it is the first work to give detailed studies based on NAS for the FAS task. Moreover, different from quick adaptation based meta NAS methods, we are the first to explore domain/type-aware meta NAS technique, which intends to find generalized architectures based on the shifted knowledge among multiple domains and attack types.

\noindent\textbf{Static-Dynamic Face Anti-Spoofing.}\quad      
In recent years, face anti-spoofing algorithms have seen great progress. Most traditional algorithms focus on handcrafted features, such as LBP~\cite{boulkenafet2015face,Pereira2012LBP}, 
SIFT~\cite{Patel2016Secure}, SURF~\cite{Boulkenafet2017Face_SURF} and HOG~\cite{Komulainen2014Context}. Other works also
focus on temporal features such as dynamic texture~\cite{komulainen2012face}, micro-motion~\cite{siddiqui2016face} and eye blinking~\cite{Pan2007Eyeblink}. More recently, a few deep learning based methods are proposed for both frame and video level liveness detection. Most works\cite{Li2017An,Patel2016Cross,george2019deep,jourabloo2018face} treat FAS as a binary classification supervised by simple binary cross-entropy loss. In contrast, pseudo depth labels~\cite{Atoum2018Face,Liu2018Learning}, reflection maps~\cite{yu2020face,kim2019basn}, and binary mask label~\cite{george2019deep} are utilized as auxiliary supervision signals as the pixel-wise guidance is able to learn more detailed information. On the other hand, according to the dynamic discrepancy~\cite{wang2018exploiting,li2019replayed}  between  live  and  spoofing faces, several video level methods are presented to exploit the dynamic spatio-temporal~\cite{wang2018exploiting,wang2020deep,yang2019face,lin2018live} or rPPG~\cite{li2016generalized,Liu2018Learning,lin2019face} features for PAD. 

Even though multi-frame dynamic methods~\cite{wang2018exploiting,Liu2018Learning} are more robust than single-frame static ones, they require more communication bandwidth and memory in terms of deployment. Inspired by the rank pooling~\cite{fernando2016rank} based dynamic representation, we propose a compact static-dynamic representation for FAS task without extra inference cost.  

\noindent\textbf{Open-Set Face Anti-Spoofing.}\quad      
Most existing FAS methods are supervised by predefined scenarios and PAs. Thus, the trained models are easy to overfit several common domains and attacks, which are vulnerable to domain shift and unseen attacks. Adversarial learning~\cite{shao2019multi}, fine-grained meta learning~\cite{shao2019regularized} and  multi-domain disentangled learning~\cite{wang2020cross} are utilized to learn robust features for domain generalization in FAS. In order to detect unseen attacks successfully, one class SVM~\cite{arashloo2017anomaly}, deep tree network~\cite{liu2019deep} and adaptive inner-update meta learning~\cite{qin2019learning} are developed.  

Despite enhancing the generalization capacity via learning strategies~\cite{shao2019multi,shao2019regularized,qin2019learning,wang2020cross}, they are still hard to explicitly learn detailed intrinsic spoofing patterns. Moreover, the existing works focus more on the learning strategies but neglect the role of architectures. In this paper, we would search well-suited networks which are able to represent discriminative and generalizable spoofing patterns (e.g., lattice artifacts) for FAS. 

\noindent\textbf{Convolution and Pooling Operators.}\quad  
In modern deep learning framework, convolution and pooling operators are the fundamental operators for feature aggregation. Recently some works extend the vanilla convolution and pooling operators to advanced version for particular applications (e.g., object detection\cite{dai2017deformable} and segmentation\cite{yu2015multi}). In terms of convolution operators, classical local descriptors (e.g., LBP \cite{ahonen2006face} and Gabor filters \cite{jain1991unsupervised}) are considered into convolution design. Representative works include Local Binary Convolution \cite{juefei2017local} and Gabor Convolution \cite{luan2018gabor}, which are proposed for saving computational cost and enhancing the resistance to the spatial changes, respectively. Besides, self-attention layer\cite{parmar2019stand} and local relation layer\cite{hu2019local} are designed for mining the local relationship flexibly. In other side, compared with vanilla average and max pooling, local importance-based pooling\cite{gao2019lip} could  automatically enhance discriminative features during the downsampling procedure. 

However, existing convolution\cite{juefei2017local,luan2018gabor,parmar2019stand} and pooling\cite{gao2019lip} operators may not be suitable for FAS task because of the limited representation capacity for intrinsic spoofing features. In order to learn robust features for domain shift as well as discriminative patterns for liveness detection, we propose central difference convolution and pooling, and develop new search space with these operators. 

\section{Static-Dynamic Central Difference Networks} \label{sec:method}
In this section we first introduce CDC and CDP in Sec.~\ref{sec:CDC} and~\ref{sec:CDP}, respectively. Based on these two operators, we propose task-aware central difference networks in Sec.~\ref{sec:CDCN}. Finally we present the static-dynamic representation in Sec.~\ref{sec:STR}. All these elements are considered in NAS-FAS.

\subsection{Central Difference Convolution} \label{sec:CDC}




The vanilla 2D convolution is the basic operator in CNNs, which consists of two main steps: 1) \textsl{sampling} local neighbor region $\mathcal{R}$ over the input feature map $x$; and then 2) \textsl{aggregating} the sampled values via learnable weights $w$. As a result, the output feature map $y$ can be formulated as

\begin{equation} 
y(p_0)=\sum_{p_n\in \mathcal{R}}w(p_n)\cdot x(p_0+p_n),
\label{eq:vanilla}
\end{equation}
where $p_0$ denotes the current location on both input and output feature maps while $p_n$ enumerates the locations in $\mathcal{R}$. For instance, local receptive field region for convolution operator with 3$\times$3 kernel and dilation 1 is $\mathcal{R}=\left \{  (-1,-1),(-1,0),\cdots,(0,1),(1,1)  \right \}$.



\begin{figure}
\centering
\includegraphics[width=7.1cm,height=3.0cm]{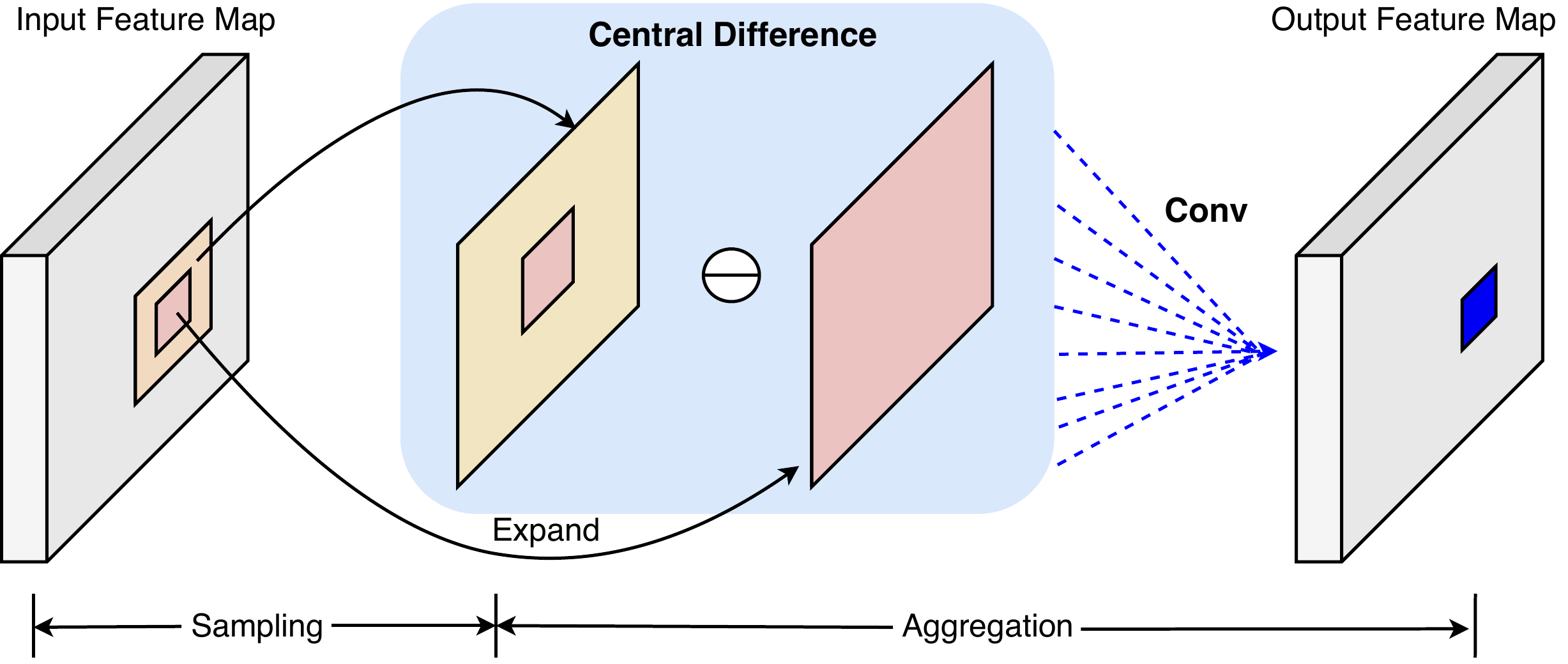}
  \caption{
  Central difference convolution. If the `Conv' is replaced by `average pooling' in the aggregation stage, it turns into central difference pooling.  
  }
\label{fig:CDC}
\end{figure}

From Eq.~\ref{eq:vanilla} we can find that vanilla convolution propagates local cues with a weighted summation manner naively, which would smooth the detailed information, and easily be influenced by sharp absolute values. 
Inspired by the LBP~\cite{boulkenafet2015face} describing local relations in a central difference way, we introduce central difference into vanilla convolution to enhance its representation and generalization capacity. As illustrated in Fig.~\ref{fig:CDC}, after sampling the local receptive field region, central difference convolution prefers to aggregate the center-oriented gradient of sampled values. Mathematically, central difference convolution is formulated as
\begin{equation} 
y(p_0)=\sum_{p_n\in \mathcal{R}}w(p_n)\cdot (x(p_0+p_n)-x(p_0)).
\label{eq:central}
\end{equation}

For FAS task, both the intensity-level semantic information and gradient-level detailed message are crucial for liveness detection, which indicates that combining vanilla convolution with central difference can be a more feasible manner to provide more robust modeling capacity. Therefore, we generalize CDC operator as 

\vspace{-1.2em}
\begin{equation} 
\begin{split}
y(p_0)
=\theta \cdot \underbrace{\sum_{p_n\in \mathcal{R}}w(p_n)\cdot (x(p_0+p_n)-x(p_0))}_{\text{central difference convolution}}&\\
+ (1-\theta)\cdot \underbrace{\sum_{p_n\in \mathcal{R}}w(p_n)\cdot x(p_0+p_n)}_{\text{vanilla convolution}},& \\
\end{split}
\label{eq:CDC}
\end{equation}
where hyperparameter $\theta \in [0,1]$ trade-offs the contribution between intensity-level and gradient-level information. The higher value of $\theta$ means the more importance of central difference gradient information. Please note that $w(p_n)$ is shared between vanilla convolution and central difference convolution, thus no extra parameters are added. The generalized \textbf{C}entral \textbf{D}ifference \textbf{C}onvolution will be referred as \textbf{CDC} henceforth. 

\noindent\textbf{Relation to Prior Work.}\quad  Here we discuss the relationship among the CDC and the existing convolutions. We also give detailed experimental comparisons in Section~\ref{sec:cdccdp}.


\textsl{Relation to Vanilla Convolution.} The CDC could be regarded as a generalized version of vanilla convolution. When $\theta$=0, the CDC degrades to vanilla convolution. In this case, the operator only aggregates local intensity information without gradient message.  

\textsl{Relation to Local Binary Convolution}\cite{juefei2017local}. Despite considering the central difference cues, local binary convolution (LBConv) utilizes fixed filters for local feature aggregation while the these filters are learnable and data-driven in the CDC. Moreover, the sparsity mechanism in LBConv limits the representation capacity.


\textsl{Relation to Gabor Convolution}\cite{luan2018gabor}. Gabor convolution (GaborConv) leverages multi-scale orientation and scale changes to enhance the robustness of spatial transformations while the CDC is good at representing detailed intrinsic features in diverse scenarios.  


\textsl{Relation to Self-Attention layer}\cite{parmar2019stand}. Self-attention models learnable relations among the local candidates, while central difference, utilized in the CDC, is one special case of various local relations. Compared with self-attention, the CDC considers the efficient FAS task-aware prior knowledge~\cite{boulkenafet2015face}, aiming to fully exploit gradient-based detailed and invariant patterns. In contrast, self-attention easily captures arbitrary relations, and learns more semantic face-aware but spoofing-unrelated features.

\subsection{Central Difference Pooling} \label{sec:CDP}
Similar to the convolution operator with two main steps (i.e., sampling and aggregation), pooling operator samples and aggregates the activation among the local receptive field region $\mathcal{R}$. Over the input feature map $x$, the average pooling operator (max pooling is analogous) can be formulated as

\begin{equation} 
y(p_0)=\frac{1}{N}\sum_{p_n\in \mathcal{R}}x(p_0+p_n),
\label{eq:avgpool}
\end{equation}
where $N$ denotes the total number of elements in local region $\mathcal{R}$. However, average pooling considers same importance for all activation, which harms discriminative intrinsic spoofing features and cause blurry downsampled features. Therefore, in order to enhance the invariant detailed representation ability, central difference gradient clues are introduced:
\vspace{-0.7em}
\begin{equation} 
y(p_0)=\frac{1}{N}\sum_{p_n\in \mathcal{R}}(x(p_0+p_n)-x(p_0)).
\label{eq:cdpool}
\end{equation}
Similar to Eq.~(\ref{eq:CDC}), we generalize the traditional intensity-level based average aggregation with central difference gradient information, which will be denoted as \textbf{C}entral \textbf{D}ifference \textbf{P}ooling (\textbf{CDP}). CDP can be formulated as 
\vspace{-0.4em}
\begin{equation} 
\begin{split}
y(p_0)
=\lambda \cdot \underbrace{\frac{1}{N}\sum_{p_n\in \mathcal{R}}(x(p_0+p_n)-x(p_0))}_{\text{central difference pooling}}&\\
+ (1-\lambda)\cdot \underbrace{\frac{1}{N}\sum_{p_n\in \mathcal{R}}x(p_0+p_n)}_{\text{average pooling}},& \\
\end{split}
\label{eq:CDP}
\end{equation}
where hyperparameter $\lambda \in [0,1]$ adjusts the contribution between intensity-level and gradient-level information, which is similar to the $\theta$ in CDC.


\noindent\textbf{Relation to Average and Max Pooling.}\quad  
Average pooling associates features with the same importance to all locations during aggregation in a small window, while max pooling only focuses on the largest activation within a neighborhood. We argue that both of them are suboptimal for FAS task. On one hand, average pooling harms discriminative and fine-grained features, which are vital for distinguishing live from spoofing faces. On the other hand, max pooling assumes that maximum
activation stands for the most discriminative detail, which is not always matched in FAS, especially when domain shifts. 

It can be seen from Eq.~(\ref{eq:CDP}) that average pooling is a special case of CDP when $\lambda=0$. For FAS task, the ‘discriminative’ and ‘robust’ features indicate fine-grained live/spoofing patterns and environment invariant clues, respectively. Local gradient operator (basic element in CDP), as a residual and difference term, is able to capture rich detailed patterns and not easily affected by external changes.

\newcommand{\tabincell}[2]{\begin{tabular}{@{}#1@{}}#2\end{tabular}}
\begin{table}\footnotesize 
    
\resizebox{0.5\textwidth}{!}{
\begin{tabular}{c|c|c|c}
	\hline
	Output & DepthNet~\cite{Liu2018Learning} & CDN\_CDC  & CDN\_CDP \\
	\hline 
	 $256\times 256$ & $3\times 3 \textrm{ conv}, 64$ & $3\times 3 \textbf{ CDC}, 64$  & $3\times 3 \textrm{ conv}, 64$\\
	 
	 \hline 
	    \tabincell{c}{$128\times 128$\\ (Low)}  & $\begin{bmatrix*}[l]
        3\times 3 \textrm{ conv}, 128\\ 
        3\times 3 \textrm{ conv}, 196\\ 
        3\times 3 \textrm{ conv}, 128\\ 
        3\times 3 \textrm{ max pool}
        \end{bmatrix*}$
        & $\begin{bmatrix*}[l]
        3\times 3\textbf{ CDC}, 128\\ 
        3\times 3\textbf{ CDC}, 196\\ 
        3\times 3\textbf{ CDC}, 128\\ 
        3\times 3\textrm{ max pool}
        \end{bmatrix*}$   
        & $\begin{bmatrix*}[l]
        3\times 3 \textrm{ conv}, 128\\ 
        3\times 3 \textrm{ conv}, 196\\ 
        3\times 3 \textrm{ conv}, 128\\ 
        3\times 3 \textbf{ CDP}
        \end{bmatrix*}$\\
	 
	 \hline 
	    \tabincell{c}{$64\times 64$ \\ (Mid)}
	    &$\begin{bmatrix*}[l]
        3\times 3 \textrm{ conv}, 128\\ 
        3\times 3 \textrm{ conv}, 196\\ 
        3\times 3 \textrm{ conv}, 128\\ 
        3\times 3 \textrm{ max pool}
        \end{bmatrix*}$ 
        & $\begin{bmatrix*}[l]
        3\times 3 \textbf{ CDC}, 128\\ 
        3\times 3 \textbf{ CDC}, 196\\ 
        3\times 3 \textbf{ CDC}, 128\\ 
        3\times 3 \textrm{ max pool}
        \end{bmatrix*}$  
         &$\begin{bmatrix*}[l]
        3\times 3 \textrm{ conv}, 128\\ 
        3\times 3 \textrm{ conv}, 196\\ 
        3\times 3 \textrm{ conv}, 128\\ 
        3\times 3 \textbf{ CDP}
        \end{bmatrix*}$ \\
        
	 \hline 
	    \tabincell{c}{$32\times 32$ \\ (High)}
	    & $\begin{bmatrix*}[l]
        3\times 3 \textrm{ conv}, 128\\ 
        3\times 3 \textrm{ conv}, 196\\ 
        3\times 3 \textrm{ conv}, 128\\ 
        3\times 3 \textrm{ max pool}
        \end{bmatrix*}$ 
        & $\begin{bmatrix*}[l]
        3\times 3 \textbf{ CDC}, 128\\ 
        3\times 3 \textbf{ CDC}, 196\\ 
        3\times 3 \textbf{ CDC}, 128\\ 
        3\times 3 \textrm{ max pool}
        \end{bmatrix*}$   
        & $\begin{bmatrix*}[l]
        3\times 3 \textrm{ conv}, 128\\ 
        3\times 3 \textrm{ conv}, 196\\ 
        3\times 3 \textrm{ conv}, 128\\ 
        3\times 3 \textbf{ CDP}
        \end{bmatrix*}$ \\
    
    \hline 
     $32\times 32$ & \multicolumn{3}{c}{[concat (Low, Mid, High), $384$]}  \\
    
    \hline 
	    $32\times 32$ & $\begin{bmatrix*}[l]
        3\times 3 \textrm{ conv}, 128\\ 
        3\times 3 \textrm{ conv}, 64\\ 
        3\times 3 \textrm{ conv}, 1 
        \end{bmatrix*}$ 
        & $\begin{bmatrix*}[l]
        3\times 3 \textbf{ CDC}, 128\\ 
        3\times 3 \textbf{ CDC}, 64\\ 
        3\times 3 \textbf{ CDC}, 1
        \end{bmatrix*}$   
        & $\begin{bmatrix*}[l]
        3\times 3 \textrm{ conv}, 128\\ 
        3\times 3 \textrm{ conv}, 64\\ 
        3\times 3 \textrm{ conv}, 1 
        \end{bmatrix*}$ \\
    
    \hline 
    \# params &  $2.25\times 10^6$ &  $2.25\times 10^6$ &  $2.25\times 10^6$\\
    
	\hline
\end{tabular}
}
\caption{Architectures of DepthNet and CDN. Inside the brackets are the filter sizes and feature dimensionalities. `conv', `CDC' and `CDP' suggest vanilla convolution, central difference convolution and pooling, respectively. All convolutional layers are with stride=1 and are followed by a BN-ReLU layer while pooling layers are with stride=2.}	
\label{tab:network}
\end{table}

\subsection{Central Difference Networks}
\label{sec:CDCN}

As pixel-wise supervision~\cite{Liu2018Learning,Atoum2018Face,yu2020searching} is proven to provide more fine-grained discrimination for liveness detection, in this paper we adopt depth-supervised framework and similar backbone ~\cite{Liu2018Learning}, called `DepthNet', as baseline. We also plug and play CDC and CDP into DepthNet to enhance the  feature representation capacity for estimating the facial depth map more accurately and robustly. The resultant network is named as \textbf{C}entral \textbf{D}ifference \textbf{N}etworks (CDN). Notably, DepthNet is the special case of the proposed CDN when using max pooling for downsampling (instead of CDP) and $\theta$=0 for all CDC operators.


Table~\ref{tab:network} shows the detailed architectures of CDN. To be specific, we replace all vanilla convolutions with CDC to form `CDN\_CDC'. Similarly, all max pooling operators are replaced by CDP to obtain `CDN\_CDP'. The CDN adopts the facial static/static-dynamic representation (with size $256\times256\times3$) as input, and then predicts the facial depth from the extracted multi-level fused features. In this paper, $\theta$=0.7 and $\lambda$=0.7 are utilized as the default setting. The ablation study about how $\theta$ and $\lambda$ trade-off the intensity and gradient clues will be conducted in Section~\ref{sec:cdccdp}. 


In terms of loss functions, classical mean square error (MSE) loss $\mathcal{L}_{MSE}$ is utilized for pixel-wise supervision. Furthermore, contrastive depth loss (CDL) $\mathcal{L}_{CDL}$~\cite{wang2020deep} is introduced to enforce the networks to learn more detailed features. Thus, the overall loss $L_{overall}$ can be formulated as $\mathcal{L}_{overall}=\mathcal{L}_{MSE}+\mathcal{L}_{CDL}$.


\subsection{Static-Dynamic Representation}
\label{sec:STR}

As the temporal clues of the particular structural live faces are different from that of PAs, temporal discrepancy (e.g., dynamic texture~\cite{tirunagari2015detection}, temporal depth~\cite{wang2018exploiting} and motion blurriness~\cite{li2019replayed}) might be helpful for FAS task. Here we consider rank pooling~\cite{fernando2016rank,wang2018cooperative} based dynamic image instead of optical flow for complementing static frame because of its superiority to regular optical flow~\cite{fernando2016rank,wang2017ordered}.

\begin{figure}
\centering
\includegraphics[scale=0.24]{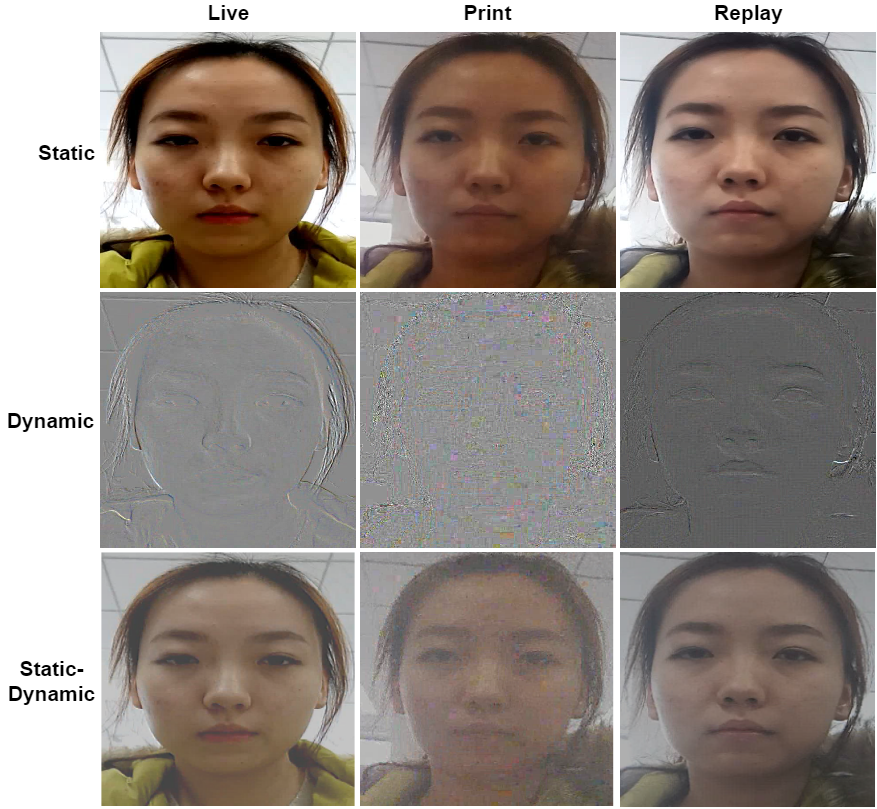}
  \caption{
  Visualization of the static, dynamic and static-dynamic live and spoofing faces.  
  }
\label{fig:STmap}
\end{figure}

\begin{figure*}
\centering
\includegraphics[width=15.0cm,height=6.5cm]{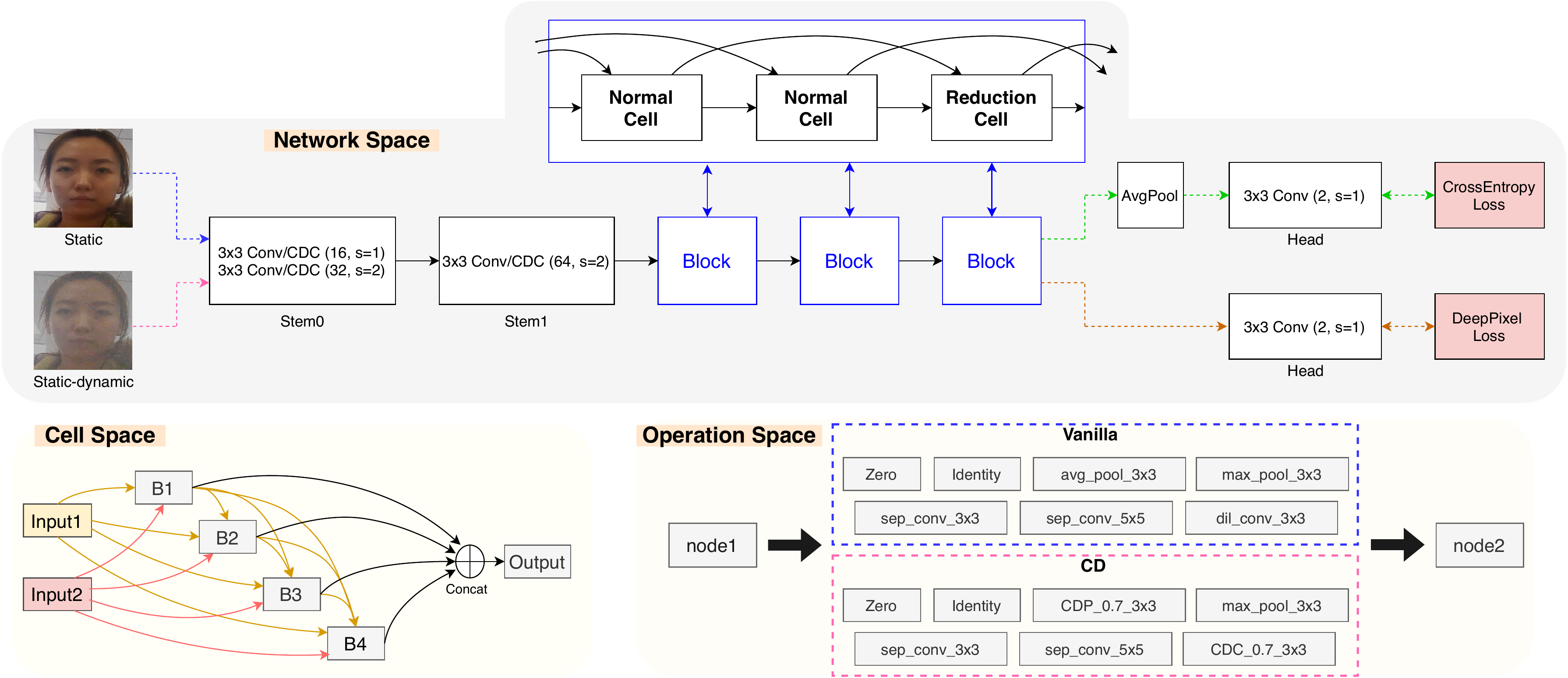}
\vspace{-0.3em}
  \caption{\small{
  Baseline search space. There are 9 layers to be searched in the network space, including six normal cells and three reduction cells. A cell contains 7 nodes, including two input nodes, four intermediate nodes B1, B2, B3, B4 and an output node. The edge between two nodes (except the output node) denotes a operation, which is chosen from the vanilla or the CD operation space.}
  }
 
\label{fig:searchspace1}
\vspace{-0.4em}
\end{figure*}

Rank pooling defines a rank function that encodes a video into a feature vector. The learning process can be seen as a convex optimization problem using the RankSVM~\cite{smola2004tutorial}. The process is formulated below
\begin{equation} 
\begin{aligned}
&\underset{D}{argmin}\frac{1}{2}\left \| D \right \|^{2}+\delta \times \sum_{i>j}\xi_{ij}\\
&  s.t. \quad D^{T}\cdot (S_{i}-S_{j})\geq 1-\xi_{ij}, \quad \xi_{ij}\geq0,
\label{eq:dynamic}
\end{aligned}
\end{equation}
where $S_{i}$ denotes the average of features over time up to $i$-frame (in sequence with $K$ frames). $\xi_{ij}$ is the slack variable, and $\delta =\frac{2}{K(K-1)}$. By optimizing Eq.~(\ref{eq:dynamic}), we map a sequence of $K$ frames to a single vector $D$. In this paper, rank pooling is directly applied on original pixels of RGB frames thus the dynamic image $D$ is of the same size as the input frame (i.e., $3 \times 256 \times 256$). In our case, given the input frame, we compute its dynamic image online with rank pooling using $K$ consecutive frames.

Despite with rich temporal information, the dynamic image always lacks detailed appearance clues, which are needed for FAS task. In order to construct a compact static-dynamic representation without extra cost for subsequent model inference, we simply add static with dynamic image and then max-min normalize it. We will also discuss other static-dynamic representation strategies in Section~\ref{sec:strepresentation}. Finally, given these static-dynamic images as input, CDN can learn discriminative and robust spatio-temporal features. 

Typical face samples are visualized in Fig.~\ref{fig:STmap}. There are obvious differences between live and spoofing faces in dynamic images despite their similarities in the original static images. It can be seen from the second row (Fig.~\ref{fig:STmap}) that the live face has more depth-aware structural clues while there are more noise patterns and lattice artifacts in the print and replay faces, respectively. After introducing dynamic patterns into static image (see the third row in Fig.~\ref{fig:STmap}), the static-dynamic representation are with sufficient appearance and temporal information for robust FAS.


\section{Neural Searching for FAS}
\label{sec:nas}

It can be seen from Table~\ref{tab:network} that the architecture of CDN is designed coarsely (e.g., simply repeating the same block structure for different levels), which might be sub-optimized for FAS task. In this section, we will briefly introduce the differentiable NAS methods\cite{liu2018darts,xu2019pc} and then present task-aware search spaces for robust searching. Finally, Domain/Type-aware Meta-NAS is introduced for efficiently searching on multiple source domains/types.


\subsection{Differentiable Architecture Search}
\label{sec:searchmethod}
In this paper, our target network to be searched is a cascade of several cells, and each cell is a directed acyclic graph (DAG) containing $N$ nodes. Each node of the graph is formed using a feature $x^{(i)}$. The edge which connects node $x^{(i)}$ and $x^{(j)}$ is denoted as $(i, j)$, and on this edge, $x^{(i)}$ passes forward to node $x^{(j)}$ through operation $f^{(i,j)}$.
Node $x^{(j)}$ is a summation of all the forward results of pre-nodes. 
Therefore, node $x^{(j)}$ can be presented as 
\begin{equation}
    x_{j}=\sum_{i}{f}^{(i,j)}(x_{i}),
	\label{eq:NAS1}
\end{equation}
where $0 \leq i < j \leq N-1$ (specifically $i=j-1$ when using FAS search space).
The operation $f^{(i,j)}$ is a composition of several operator candidates (convolution, pooling \emph{etc.}). So the operation $f^{(i,j)}$ can be represented as 
\begin{equation}
\begin{split}
    f^{(i,j)}(x_i)=\sum_{o\in \mathcal{O}} \beta_{o}^{(i,j)} \cdot o(x_{i}), \\
    \beta_{o}^{(i,j)} = \frac{exp(\alpha_{o}^{(i,j)})}{\sum_{{o}'\in \mathcal{O}}exp(\alpha_{{o}'}^{(i,j)})},
\end{split}
	\label{eq:NAS_edge_function}
\end{equation}
where $\mathcal{O}$ is the set of candidate operations, and $o(x_{i})$ is the output of operation $o$ with node $x_i$ as the input.
$\beta_{o}^{(i,j)}$ is the weight of the operation $o$ in $f^{(i,j)}(x_i)$, and when $\beta_{o}^{(i,j)}$ getting larger and larger, $f^{(i,j)}$ is more and more determined by the operation $o$.
$\alpha_{o}^{(i,j)}$ is a trainable variable that is used to calculate $\beta_{o}^{(i,j)}$ with the softmax function.
As a summary, all the trainable $\alpha_{o}^{(i,j)}$ determines the network architecture.
So, the task of searching the network architecture turns to optimizing all $\alpha_{o}^{(i,j)}$ in the network.

\begin{figure*}
\centering

\includegraphics[scale=0.31]{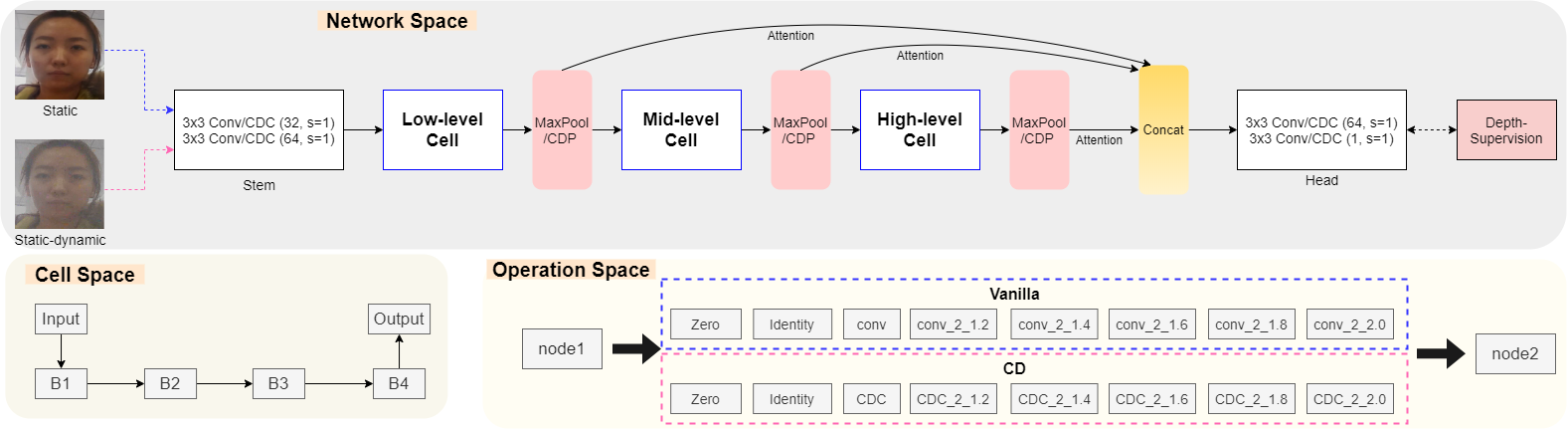}
\vspace{-0.4em}
  \caption{\small{
  FAS search space. There are three cells (one low-level, one mid-level and one high-level) to be searched.} 
  }
\label{fig:searchspace2}
\vspace{-0.4em}
\end{figure*}

\noindent\textbf{Optimization.}\quad  Following the similar bi-level optimization strategy \cite{liu2018darts,xu2019pc}, we denote $\alpha = \{\alpha_{o}^{(i,j)} \}$ as the set of all $\alpha_{o}^{(i,j)}$, and $\alpha$ presents the network architecture.
Network weight $\varphi$ and architecture $\alpha$ are optimized alternatively on the support and query sets.  
On the support set, the optimization of $\varphi$ can be formulated as
\begin{equation}
	\varphi(\alpha) = \varphi - \gamma_1 {\boldmath \cdot} \nabla_{\varphi} \mathcal{L}_{s}(\varphi, \alpha),
	\label{eq:optimizing of theta}
\end{equation}
where $\varphi(\alpha)$ is the update result of $\varphi$ conditioned by current architecture $\alpha$.
$\gamma_1$ and $\mathcal{L}_{s}$ are the learning rate and loss on the support set, respectively.
On the query set, the optimization of $\alpha$ can be formulated as
\begin{equation}
\begin{split}
    \alpha =& \alpha - \gamma_2 {\boldmath \cdot} \nabla_{\alpha} \mathcal{L}_{q}(\varphi(\alpha), \alpha)\\
     =& \alpha - \gamma_2 {\boldmath \cdot} \nabla_{\alpha} \mathcal{L}_{q}(\varphi - \gamma_1 {\boldmath \cdot} \nabla_{\varphi} \mathcal{L}_{s}(\varphi, \alpha), \alpha),
\end{split}
	\label{eq:optimizing of alpha}
\end{equation}
where $\gamma_2$ and $\mathcal{L}_{q}$ are the learning rate and the loss on the query set, respectively.
By alternatively optimizing $\varphi$ and $\alpha$ with Eq.(\ref{eq:optimizing of theta}) and Eq.(\ref{eq:optimizing of alpha}), the searching stage converges gradually. After searching, the operations with the largest weight $max_{o\in \mathcal{O},o\neq none}\,\beta_o^{(i,j)}$ and $M$ incoming edges with $M$ largest $max_{o\in \mathcal{O},o\neq none}\,\beta_o^{(i,j)}$ are adopted to form the final discrete architecture ($M=2$ for baseline search space while $M=1$ for FAS search space).

\subsection{Baseline Search Space}
\label{sec:searchspace1}
The search space covers all possible candidate CNN to be found, and is important for NAS. A standard search space in NAS is `NASNet search space'\cite{zoph2018learning}. Here we establish similar baseline search space, which is illustrated in Fig.~\ref{fig:searchspace1}.


\noindent\textbf{Network Space.}\quad The network space is comprised of 12 layers (two stem, nine cell and one head layers). We search for two kinds of cells in networks, i.e., normal and reduction cells. For the normal cell, each operator has the stride of 1 while the first operator has the stride of 2 for the reduction cell. The input nodes of each cell are propagated from the output nodes of two previous cells. 

In terms of loss function, the network space takes $3 \times 256 \times 256$ static or static-dynamic image as input and predict a scalar score or $8 \times 8$ binary map. The former one treats FAS as binary classification task supervised by common cross-entropy loss while the latter utilizes pixel-wise binary (DeepPixel) loss~\cite{george2019deep}. 

\noindent\textbf{Cell Space.}\quad Each cell contains seven nodes, including two input nodes, four intermediate nodes and one output node. The edge connections to the intermediate nodes denote summation operation while the output node concatenates all results from intermediate nodes. 

\noindent\textbf{Operation Space.}\quad There are two kinds of operation spaces (i.e., vanilla and central difference (CD)) in our setting. As shown in Fig.~\ref{fig:searchspace1}, they share most operator candidates but CD space utilizes `CDP\_0.7\_3x3' and `CDC\_0.7\_3x3' instead of 'avg\_pool\_3x3' and 'dil\_conv\_3x3', respectively. The total search space is $(7^{(1+2+3+4)})^{2}=7^{20}$.   

In summary, there are eight trials to be explored for baseline search space, including `static vanilla space with cross-entropy loss (S-Van-CE)', `static-dynamic CD space with DeepPixel loss (SD-CD-DP)', `S-Van-DP', `SD-Van-CE', `SD-Van-DP', `S-CD-CE', `S-CD-DP' and `SD-CD-CE'. The corresponding ablation study is conducted in Section~\ref{sec:nasbaseline}.

\subsection{FAS Search Space}
\label{sec:searchspace2}

As mentioned in Section~\ref{sec:CDCN}, DepthNet~\cite{Liu2018Learning} performs well in FAS task because of exploiting multi-level fused features and fine-grained depth-supervision. Based on the task-aware knowledge, we establish novel FAS search spaces, which is illustrated in Fig.~\ref{fig:searchspace2}.

\noindent\textbf{Network Space.}\quad Inspired by the structure of DepthNet, our network space consists of one stem and head layers and low-mid-high level cells. There is a pooling layer (max pooling or CDP) with or without spatial attention~\cite{woo2018cbam} after each cell. The attention module forces the cells to learn more concentrated features, which is proved to be effective~\cite{wang2019multi} for FAS task. Finally, the low-mid-high level features are concatenated for prediction. As for loss function, we utilize $\mathcal{L}_{s}=\mathcal{L}_{q}=\mathcal{L}_{MSE}+\mathcal{L}_{CDL}$ (same as CDN in Section~\ref{sec:CDCN}). 

\noindent\textbf{Cell Space.}\quad Each cell contains six nodes, including one input node, four intermediate nodes and one output node. The edge only connects to the adjacent nodes while the output node adopts the result from the last intermediate node directly. 

\noindent\textbf{Operation Space.}\quad There are also two kinds of operation spaces (i.e., vanilla and central difference (CD)) in our setting. As shown in Fig.~\ref{fig:searchspace2}, the CD operation space utilizes `CDC' and `CDC\_2\_$r$' instead of 'conv' and 'conv\_2\_$r$', respectively. All the convolution operators are with $3 \times 3$ kernels and $\theta=0.7$ for all CDC. And `\_2\_$r$' means using two stacked convolutions to increase channel number with ratio $r$ first and then decrease back to the original channel size. The total search space is $(8^{4})^{3}=8^{12}$.   

Overall, there are 10 trials to be explored for FAS search space, including `static vanilla space with max pooling without attention (S-Van-Max)', `static-dynamic CD space with CDP with attention (SD-CD-CDP-Att)', `S-CD-Max', `S-CD-CDP', `S-CD-Max-Att', `S-CD-CDP-Att', `SD-Van-Max', `SD-CD-Max', `SD-CD-CDP' and `SD-CD-Max-Att'. The ablation study will be shown in Section~\ref{sec:nasFAS}.

\begin{figure*}
\includegraphics[scale=0.47]{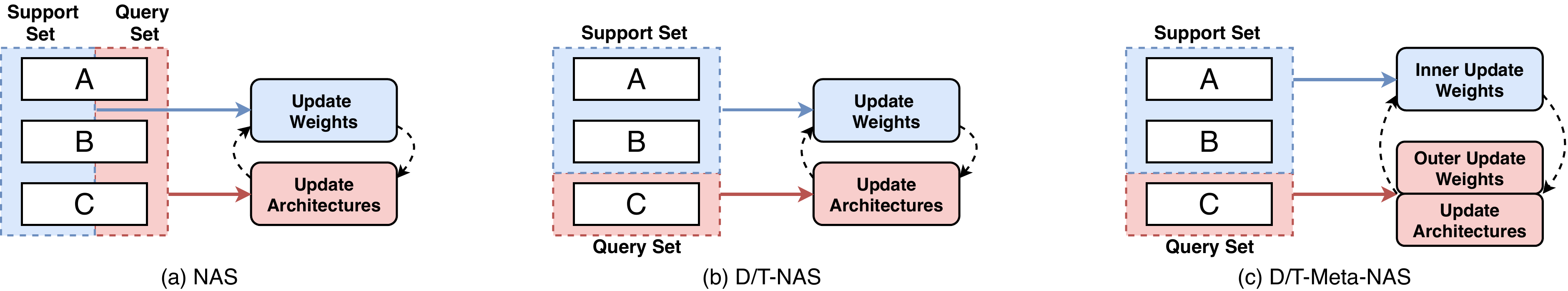}
 \vspace{-0.8em}
  \caption{
  Neural searching on the datasets with multiple domains or attack types. We use DARTS based NAS for example. The 'A', 'B' and 'C' denote the data from three respective shifted domains or unseen attack types, which can be easily extended to larger ($\textgreater$ 3) cases in practice. (a) NAS: randomly sample the support and query sets from entire data, and then search with bi-level optimization strategy. (b) D/T-NAS: first divide the support set and query set according to the prior knowledge of domains or attack types, and then search. (c) D/T-Meta-NAS: first meta-train the network weights with domain/type-shifted knowledge, and then update the architecture.}
 \vspace{-0.8em}
\label{fig:MetaNAS}
\end{figure*}

\subsection{Domain/Type-aware Meta-NAS}
\label{sec:meta}

Although the existing NAS methods are able to search well-suited architectures in a given dataset (domain), it is still unknown how NAS performs when searching in multiple source domains. As for FAS task, it is practical to collect data from various scenarios (e.g., environment and attack types) for searching and training. It is valuable if the searched networks from multiple given source domains or attack types could generalize well in unseen domain/type.


Here we propose Domain/Type-aware Meta-NAS (D/T-Meta-NAS), which leverages the prior domain/type knowledge for better searching. The complete algorithm of D/T-Meta-NAS is summarized in Algorithm~\ref{algorithm:Meta-FAS-BS}. 
Suppose that we have access to $N$ source domains/types of FAS task $D=[D_1,D_2,...,D_N]$, in which the randomly selected $N-1$ domains/types are chosen as the support domains $D^{s}$ while the rest one as the query domain $D^{q}$. Then we sample batch examples $\mathcal{B}_{i}^{s}(i=1,2,...,N-1)$ in every domain/type of $D^{s}$, and $\mathcal{B}^{q}$ in $D^{q}$, which are used for inner-updated and optimization stage, respectively.


\noindent\textbf{Inner-Updated Stage.}\quad In this stage, the meta-learner with weight $\varphi$ inner-updates itself on each $\mathcal{B}_{i}^{s}$. For simplicity, we show only one inner-update step, which can be formulated as 
\begin{equation}
\varphi_{i}(\alpha) = \varphi - \gamma_1 {\boldmath \cdot} \nabla_{\varphi} \mathcal{L}_{s}(\mathcal{B}_{i}^{s},\varphi, \alpha), \quad i=1,...,N-1, 
	\label{eq:inner}
\end{equation}
where $\varphi_{i}(\alpha)$ is the meta-learner’s updated weight on the $i-th$ batch data $\mathcal{B}_{i}^{s}$. 
$\gamma_1$ denotes the learning rate of the meta-learner at the inner-updated stage, and $\mathcal{L}_{s}(\mathcal{B}_{i}^{s},\varphi, \alpha)$ is the meta-learner’s loss on the batch data $\mathcal{B}_{i}^{s}$ with respect to architecture $\alpha$. 
After inner-update, the meta-learner turns to $N-1$ domain-specific learners with weights $\varphi_{i}(\alpha)$ where $i=1,...,N-1$.


\begin{algorithm}[t]
	\caption{D/T-Meta-NAS}
	{\bfseries Input:} Training data $D$ with $N$ domains/types, learning rates $\gamma_1,\widetilde{\gamma}_1,\gamma_2$ \\
	{\bfseries 1\,\,\,:} Initialize meta-learner weight $\varphi$ and architecture $\alpha$ \\
	{\bfseries 2\,\,\,:} {\bfseries while } not done {\bfseries do} \\
	{\bfseries 3\,\,\,:}  \ \, Randomly select $N$-1 domains/types in $D$ as support domains $D^{s}$, and the remaining one as query domain $D^{q}$ \\
	{\bfseries 4\,\,\,:}  \ \ \,  Sample batch examples $\mathcal{B}_{i}^{s}(i=1,...,N-1)$ in every domain/type of $D^{s}$, and examples $\mathcal{B}^{q}$ in $D^{q}$ \\
	{\bfseries 5\,\,\,:}  \ \ \,   {\bfseries for} each  $\mathcal{B}_{i}^{s}$  {\bfseries do} \\
	{\bfseries 6\,\,\,:}  \ \ \quad \,  $\varphi_{i}(\alpha) = \varphi - \gamma_1 {\boldmath \cdot} \nabla_{\varphi} \mathcal{L}_{s}(\mathcal{B}_{i}^{s},\varphi, \alpha) $  \\
	{\bfseries 7\,\,\,:} 	\ \ \, {\bfseries end} \\
	{\bfseries 8\,\,\,:}  \ \ \,  $\varphi(\alpha) = \varphi - \widetilde{\gamma}_1 {\boldmath \cdot} \nabla_{\varphi} \sum_{i}^{N-1}\mathcal{L}_{q}(\mathcal{B}^{q},\varphi_{i}(\alpha), \alpha) $  \\
	{\bfseries 9\,\,\,:}  \ \ \,  $\alpha = \alpha - \gamma_2 {\boldmath \cdot} \nabla_{\alpha} \mathcal{L}_{q}(\mathcal{B}^{q}, \varphi(\alpha), \alpha)$ \\
	{\bfseries 10:} \ \ \, $\varphi = \varphi(\alpha)$  \\
	{\bfseries 11:} 	{\bfseries end while} \\
	{\bfseries 12:} 	{\bfseries return} architecture $\alpha$
	\label{algorithm:Meta-FAS-BS}
\end{algorithm}

\noindent\textbf{Optimization Stage.}\quad Each learner with weight $\varphi_{i}(\alpha)$ is evaluated on the query data $B^{q}$, which contains face images belonging to unseen query domain/type. 
Then, the meta-learner is optimized using all learners' loss on the query data $B^{q}$.
With outer-updated $\varphi(\alpha)$, architecture $\alpha$ is subsequently optimized on $B^{q}$. The whole optimization can be formulated as
 \vspace{-0.8em}
\begin{equation}
\varphi(\alpha) = \varphi - \widetilde{\gamma}_1 {\boldmath \cdot} \nabla_{\varphi} \sum_{i}^{N-1}\mathcal{L}_{q}(\mathcal{B}^{q},\varphi_{i}(\alpha), \alpha),
	\label{eq:outer1}
\end{equation}

\begin{equation}
\alpha = \alpha - \gamma_2 {\boldmath \cdot} \nabla_{\alpha} \mathcal{L}_{q}(\mathcal{B}^{q}, \varphi(\alpha), \alpha),
	\label{eq:outer2}
\end{equation}
where $\mathcal{L}_{q}(\mathcal{B}^{q},\varphi_{i}(\alpha), \alpha)$ and $\mathcal{L}_{q}(\mathcal{B}^{q}, \varphi(\alpha),\alpha)$ are the $i-th$ learner’s loss with respect to architecture $\alpha$, and architecture's loss with respect to the weights $\varphi(\alpha)$ on $\mathcal{B}^{q}$, respectively.
$\widetilde{\gamma}_1$ and $\gamma_2$ denote the learning rate of meta-learner and architecture in the optimization stage, respectively. 
Note that, in Eq.~\ref{eq:outer1}, $\nabla_{\varphi} \sum_{i}^{N-1}\mathcal{L}_{q}(\mathcal{B}^{q},\varphi_{i}(\alpha), \alpha)$ uses the learners' losses on the query data $\mathcal{B}^{q}$ to compute the gradient of $\varphi$, but not $\varphi_{i}(\alpha)$. After obtaining the stable updated meta-weights $\widetilde{\gamma}_1$, architecture $\alpha$ is then updated according to the loss $\mathcal{L}_{q}(\mathcal{B}^{q},\varphi(\alpha),\alpha)$ via Eq.~\ref{eq:outer2}.  


By iteratively meta-training weights and updating architectures on the domain/type-aware tasks, the meta-learner learns $\varphi$ towards right directions based on domain/type shifted knowledge among different domains while $\alpha$ is subsequently updated robustly due to the reliable $\varphi$. In other words, with the meta-learned weights, the searched architecture is more likely to detect the spoofing faces with unseen domains/types by efficiently learning and searching on the support set with its learned preferable adaptive inner-update rule.

\noindent\textbf{Discussion.}\quad Here we give comparisons with the schemes of NAS, Domain/Type-aware NAS (D/T-NAS), and the proposed D/T-Meta-NAS, which are illustrated in Fig.~\ref{fig:MetaNAS}. The corresponding ablation study on cross-dataset intra-type protocol will be shown in Section~\ref{sec:RandomSampling}.

\textsl{NAS vs. D/T-NAS}.\quad Traditional NAS (e.g., DARTS~\cite{liu2018darts}) randomly sample half data from the training set as support set while the remaining half as query set. In contrast, D/T-NAS (Fig.~\ref{fig:MetaNAS}(b)) divides the data space with fine-grained domain/type knowledge. Specifically, randomly select $N$-1 domains/types as support set, and the remaining one as query set. In the support set, the sampling rule for tasks is also domain/type-aware. Thus, the architecture search will try to optimize towards the unseen domain/type, which has not been used in the weight-updated support set. 

\textsl{D/T-NAS vs. D/T-Meta-NAS}.\quad Similar to D/T-NAS, D/T-Meta-NAS (Fig.~\ref{fig:MetaNAS}(c)) adopts domain/type-aware partition between support set and query set, as well as for task generation. Furthermore, the weights are meta-learned from the fine-grained tasks with different domains/types, which sufficiently exploits the domain/type knowledge in support set, and mimics the unseen domain/type in query set. The architecture updates based on the domain/type generalized weights, which is more stable and not easily influenced by the domain/type shifted discrepancy.


\section{Datasets and Protocols}
\label{sec:dataset}

In this section, we first present the CASIA-SURF 3DMask dataset, and then introduce four FAS testing protocols. The CASIA-SURF 3DMask dataset is now available at \href{http://www.cbsr.ia.ac.cn/users/jwan/database/3DMask.pdf}{http://www.cbsr.ia.ac.cn/users/jwan/database/3DMask}.  

\subsection{CASIA-SURF 3DMask Dataset}
With the 3D print technology becoming more and more popular, the 3D mask attacks built by 3D printing attract attention in FAS community. However, existing 3D mask FAS datasets have various degrees of drawbacks (e.g., low video quality, small amount of subjects and videos, laboratory controlled environment and unrealistic mask appearance). As a result, the mask attacks could be easily detected even using common model (e.g., ResNet50~\cite{he2016deep}), which will be discussed in Section~\ref{sec:cross-cross}.

\begin{figure}
\includegraphics[scale=0.62]{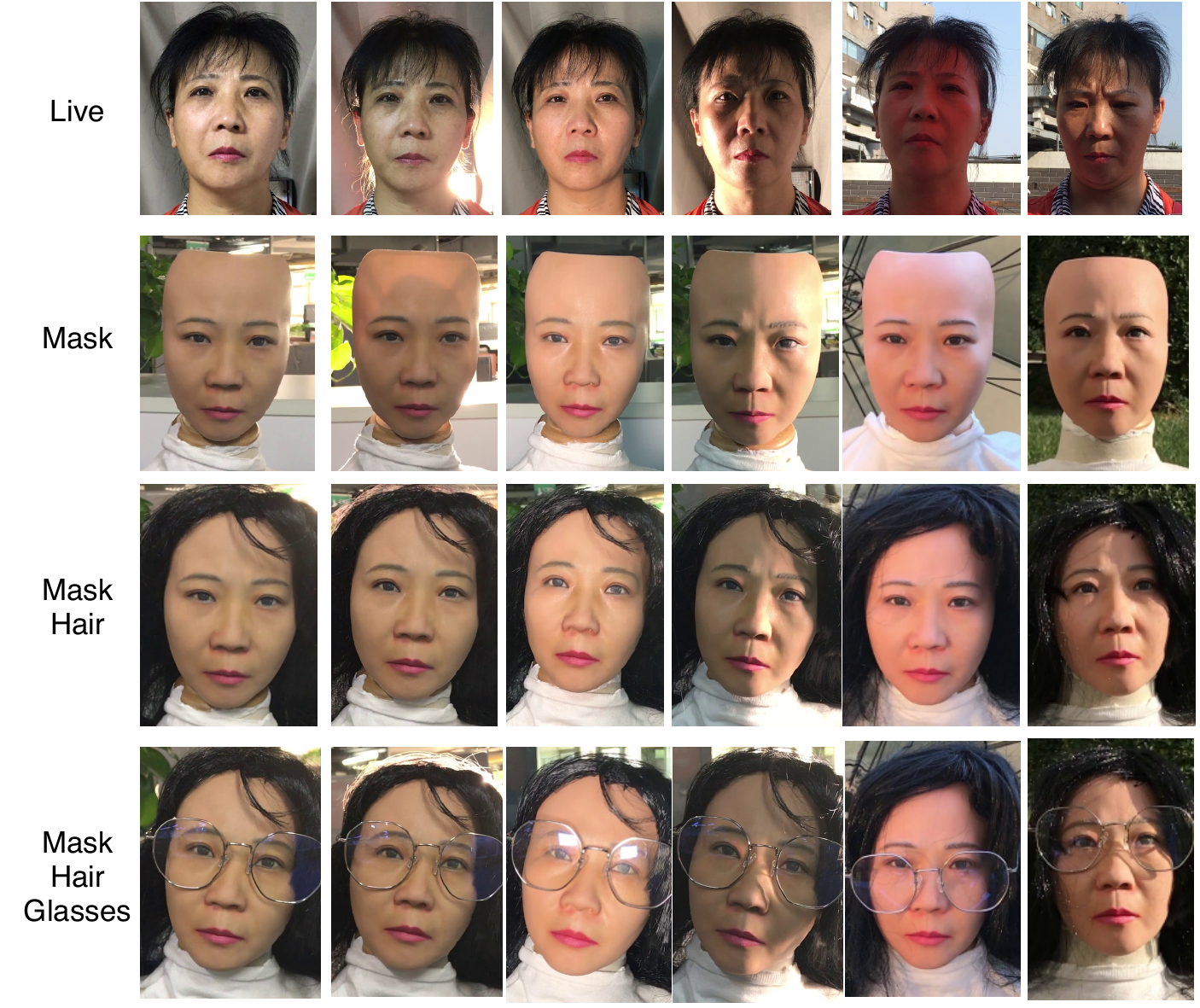}
  \caption{
  Samples of the CASIA-SURF 3DMask dataset. The left four columns are indoor while the right two ones are outdoor scenes.
  }
 
\label{fig:MASK}
\end{figure}

In order to study the generalization ability of detecting realistic 3D mask in the wild, we build up a novel large-scale 3D mask dataset CASIA-SURF 3DMask (briefly named 3DMask). For live data, we include 288 videos from 48 subjects (six videos per subject). In consideration of real-world variant environment, six conditions are adopted for data acquisition, including normal, back-light, front-light, side-light, outdoor in shadow and outdoor in sunlight. To our best knowledge, this is the first FAS dataset considering outdoor scenes with challenging lighting. For spoofing data collection, we collected 3D masks with 48 subjects via 3D printing. Besides using only the naive masks, we also consider two more realistic decoration cases (i.e., masks with/without hair and glasses). Thus totally 864 mask videos are recorded (48 subjects with three mask decorations and six environment conditions). Compared with two classical 3D mask datasets (3DMAD~\cite{erdogmus2014spoofing} and HKBU-MARs~\cite{liu20163d}), our 3DMask not only has larger number of realistic 3D masks but also considers complex mask decoration cases. 

In the 3DMask dataset, videos are captured with latest mobile devices with several brands (i.e., Apple, Huawei and Samsung). Each video sequence lasts for about 10 seconds with frame rate 30 fps and 1080p resolution. All recorded subjects are Chinese people (21 males and 27 females). In terms of the age distribution, most of the subjects are within the range [20,30) and [50,60) years old. The youngest and eldest age is 23 and 62, respectively. Some live and spoofing samples are displayed in Fig.~\ref{fig:MASK}.

\subsection{FAS Protocols}
Nine databases OULU-NPU~\cite{Boulkenafet2017OULU}, SiW~\cite{Liu2018Learning}, CASIA-MFSD~\cite{Zhang2012A}, Replay-Attack~\cite{ReplayAttack}, MSU-MFSD~\cite{wen2015face}, SiW-M~\cite{liu2019deep}, 3DMAD~\cite{erdogmus2014spoofing}, HKBU-MARs~\cite{liu20163d} and the proposed 3DMask are used in the four FAS testing protocols. The first two protocols is used to evaluate the model robustness under domain shifts while the last two protocols measures the model generalization ability to unseen attack types (especially the last protocol is with both serious domain shifts and unseen attack types).

\noindent\textbf{Intra-Dataset Intra-Type Protocol~\cite{Boulkenafet2017OULU,Liu2018Learning}.}\quad 
In training and testing stages, it uses the same dataset with the same attack types but changes acquisition conditions. The OULU-NPU and SiW datasets utilized for generalization validation. We strictly follow the four sub-protocols on OULU-NPU~\cite{Boulkenafet2017OULU} and three sub-protocols on SiW~\cite{Liu2018Learning} for fair evaluation. In terms of performance metrics, Attack Presentation Classification Error Rate (APCER), Bona Fide Presentation Classification Error Rate (BPCER), and ACER are utilized.


\noindent\textbf{Cross-Dataset Intra-Type Protocol~\cite{shao2019multi,shao2019regularized}.}\quad  
This protocol focuses on cross-dataset level domain generalization ability measurement, which usually trains models on several datasets (multiple domains) and then tests on unseen datasets (shifted domain). CASIA-MFSD, Replay-Attack, MSU-MFSD and OULU-NPU are utilized for this protocol, which follows `leave one dataset out' principle. In this protocol, Half Total Error Rate (HTER) and AUC are adopted for performance metrics. 

\noindent\textbf{Intra-Dataset Cross-Type Protocol~\cite{arashloo2017anomaly,liu2019deep}.}\quad  The protocol adopts `leave one attack type out' to validate the model robustness for unseen attack types, i.e., one kind of attack type only appears in testing stage. Considering the rich attack types, CASIA-MFSD, Replay-Attack, MSU-MFSD and SiW-M are utilized in this protocol. As for performance metrics, Area Under Curve (AUC) is utilized for first three datasets while APCER, BPCER, ACER and Equal Error Rate (EER) are employed for SiW-M. 

\noindent\textbf{Cross-Dataset Cross-Type Protocol.}\quad Although the above-mentioned three protocols mimic most factors in real-world applications, they do not consider the most challenging case, i.e., cross-dataset cross-type testing. 
In order to measure the generalization of both unseen domain and attack types, we propose the novel `cross-dataset cross-type' protocol. OULU-NPU and SiW are mixed for training while 3DMAD, HKBU-MARs and 3DMask are used for testing. As for performance metrics, AUC, ACER and HTER are utilized.

\section{Experiments}
\label{sec:experiment}

In this part, we first give details for experimental setup. Then, we thoroughly evaluate the impacts of central difference family, static-dynamic representation, baseline and FAS search space on Protocol-1~\cite{Boulkenafet2017OULU} (domain shift with illumination condition and location) of OULU-NPU. Besides, we verify the effectiveness of D/T-Meta-NAS when searching on multiple domains on cross-dataset intra-type protocol. Finally we show the state-of-the-art results of the proposed methods on nine datasets with four testing protocols.

\subsection{Implementation Details}

\noindent\textbf{Ground Truth Generation.}\quad
 The facial depth map label is generated by the off-the-shelf 3D face model~\cite{Feng2018Joint}. The binary map for DeepPixel~\cite{george2019deep} is generated simply by downsampling the face image and filling each patch position with corresponding binary label. The generated binary and depth maps keep the same size with $32\times32$. The live depth map is normalized in a range of $[0, 1]$, while the spoof one is all 0 at the training stage, which is beneficial for learning discriminative patterns for FAS task. 
 

\noindent\textbf{Training and Testing Setting.}\quad 
Rank pooling based dynamic image is generated with hyperparameter $K$=7. As for the attention module, spatial size $7\times7$, $5\times5$ and $3\times3$ are utilized for low, mid, high level, respectively. We use Pytorch framework for implementation. At the training stage, Adam optimizer with weight decay (wd=5e-5)is used. We set the initial learning rate (lr=1e-4), which halves every 500 epochs. Our models are trained with batchsize 8 on a single P100 GPU for maximum 1300 epochs. At the testing stage, the decision score is simply generated via mean pooling the predicted binary/depth map.

\noindent\textbf{Searching Setting.} \quad In order to search efficiently with less memory cost, we adopt partial channel connection and edge normalization\cite{xu2019pc}. In the searching phase, the channel numbers are according to Fig.~\ref{fig:searchspace1} and Fig.~\ref{fig:searchspace2}, which would be doubled in the retraining and testing phases. At the searching stage, Adam optimizer (with lr=1e-4 and wd=5e-5) is used for updating weights $\varphi$, while architectures $\alpha$ are updated via Adam (with lr=6e-4 and wd=1e-3). The search is conducted on Protocol-1 of OULU-NPU with batchsize 10 for 60 epochs. Notably, $\alpha$ are fixed in the first 15 epochs for stable $\varphi$ initialization. Specifically, D/T-Meta-NAS is searched on multiple domains (cross-dataset intra-type protocol) or types (intra-dataset cross-type protocol), where batchsize=8 is used for each domain/type task. Learning rate $\gamma_1$=$\widetilde{\gamma}_1$=1e-4 is utilized for D/T-Meta-NAS.


\begin{figure}
\centering
\includegraphics[scale=0.52]{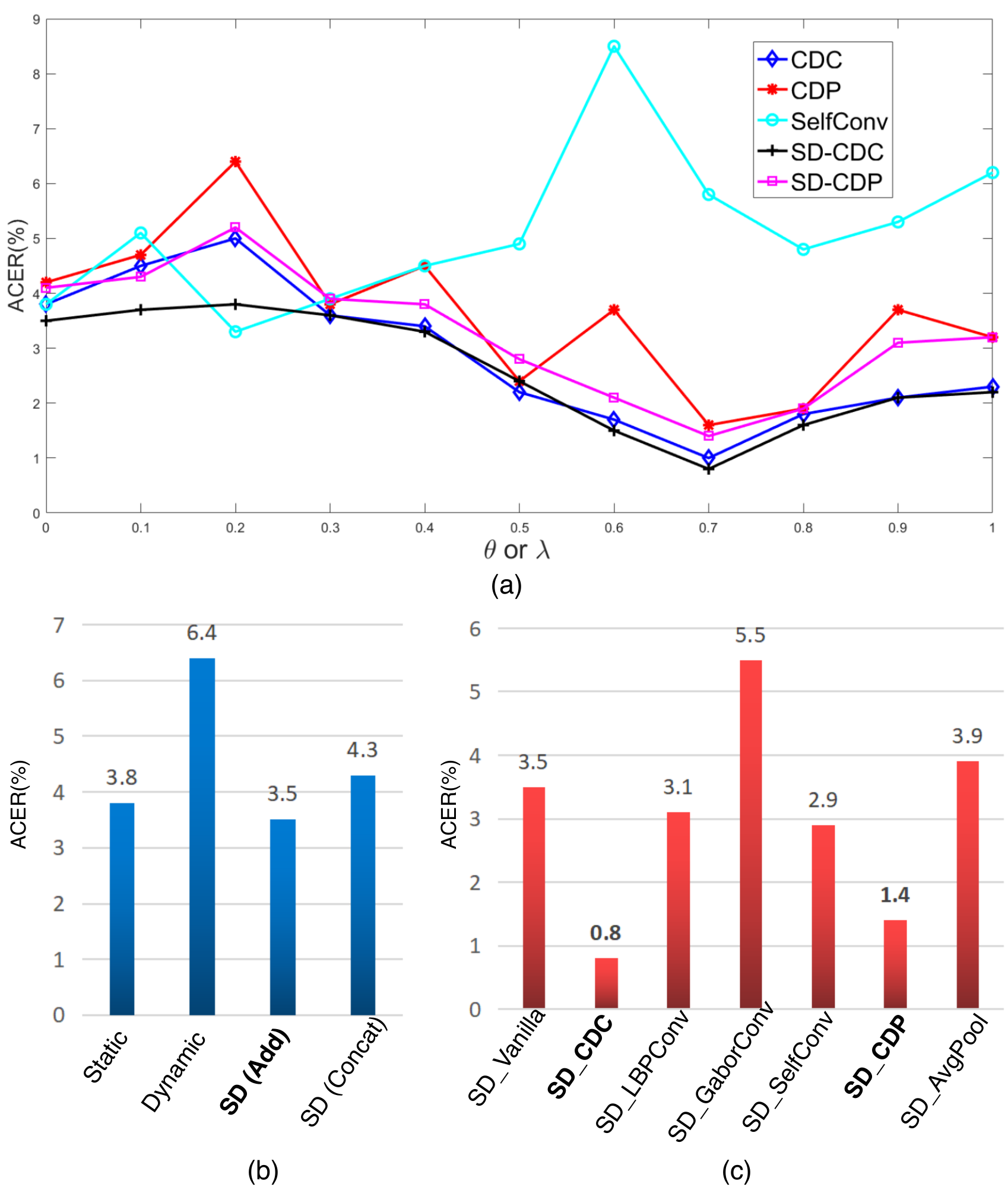}
  \caption{
  Ablation study of CDC, CDP and static-dynamic representation. (a) Impact of $\theta$ and $\lambda$ in CDN. (b) Impact of dynamic representation. (c) Comparison among various convolutions and pooling. `SD' is short for static-dynamic representation. Lower ACER indicates better performance. 
  }
 
\label{fig:Experiment1}
\end{figure}

\subsection{Impact of CDC and CDP}
\label{sec:cdccdp}

\noindent\textbf{Impact of $\theta$ and $\lambda$ in CDN.}\quad 
According to Eq.~(\ref{eq:CDC}) and Eq.~(\ref{eq:CDP}), $\theta$ and $\lambda$ control the contribution of the gradient-based details, i.e., the higher $\theta$, the more local detailed information included. As illustrated in the blue (CDC) and red (CDP) broken lines Fig.~\ref{fig:Experiment1}(a), with larger $\theta$ and $\lambda$, CDN can achieve better performance than vanilla convolution ($\theta$=0, ACER=3.8\%) and average pooling ($\lambda$=0, ACER=4.1\%), indicating the central difference based fine-grained information is helpful for FAS task. The best results could be obtained when $\theta$=0.7 and $\lambda$=0.7 for CDC and CDP, respectively. It is interesting to find that CDC and CDP also perform well for FAS task even in the extreme case $\theta$=1.0 and $\lambda$=1.0, i.e., only considering the gradient-based cues.


\noindent\textbf{CDC vs. Other Convolutions.}\quad At first, we evaluate the effectiveness of the self-attention (learnable local relation) in FAS task. We follow the structure of self-attention in~\cite{parmar2019stand} and extend it to a generalized version (like CDC), which can be formulated as $SelfConv=\theta*SelfAttention+(1-\theta)*VanillaConv$. As illustrated in Fig.~\ref{fig:Experiment1}(a), `SelfConv' is with negative effects in most settings of $\theta$ while only decreases 0.5\% ACER when $\theta$=0.2. It indicates that it is challenging to capture intrinsic spoofing patterns with arbitrary learnable local relations.

Then we give the comparisons among various convolutions for FAS task. Here all configurations are with best hyperparameters and static-dynamic inputs. The first five columns of Fig.~\ref{fig:Experiment1}(c) shows that CDC outperforms other convolutions (i.e., vanilla, LBConv~\cite{juefei2017local}, GaborConv~\cite{luan2018gabor} and SelfConv~\cite{parmar2019stand}) by a large margin (more than 2\% ACER). It is interesting to find that LBConv performs better than vanilla convolution, indicating that the local gradient information is important for FAS task. GaborConv performs the worst because it is designed for capturing spatial invariant features, which is not suitable for FAS task.  


\noindent\textbf{CDP vs. other Poolings.}\quad 
As shown in the last two columns of Fig.~\ref{fig:Experiment1}(c), CDP outperforms traditional average pooling by 2.5\% ACER because it introduces the fine-grained gradient patterns to avoid excessive local blurriness caused by average operation. CDP also achieves better performance than max pooling (see the column `SD\_Vanilla'), indicating the central diffrence clues are more discriminative to detect the spoofing attacks.

The reasons that CD family performs well are two fold: 1) Central difference gradient clues is helpful to represent the local detailed intrinsic spoofing patterns (e.g., lattice artifacts shown in Fig.~\ref{fig:Figure2}), which is discriminative for FAS task; 2) Local gradient operator (basic element in CD family), as a residual and difference term, is not easily affected by external changes (e.g., illumination), which is robust for domain shifts. In summary, both CDC and CDP hold great performance in human-designed CDN architecture for FAS task, which motivates us to consider them into search space for subsequent neural searching.

\begin{figure}
\centering
\includegraphics[scale=0.6]{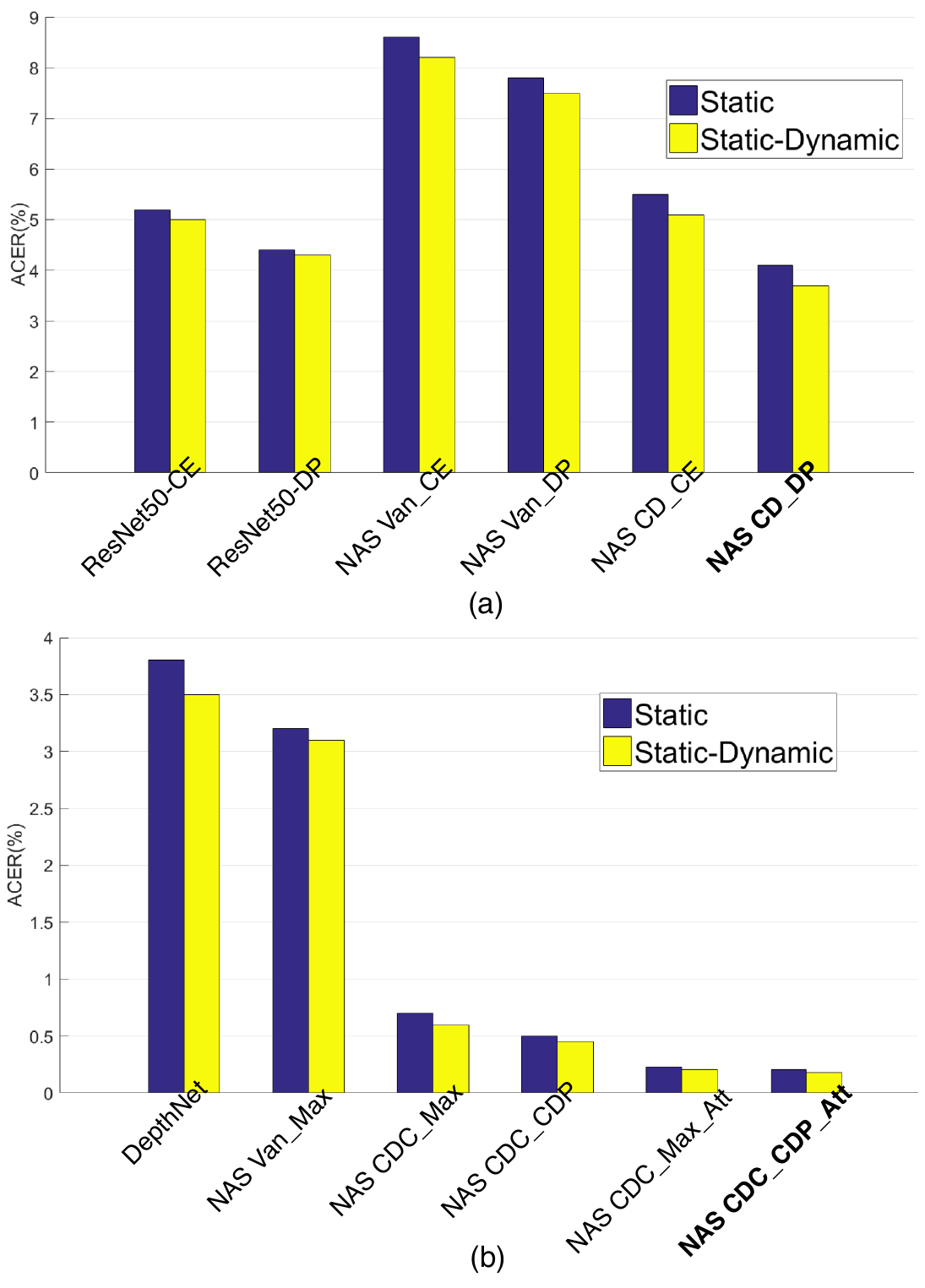}
  \vspace{-0.6em}
  \caption{
  Ablation study of search space components. (a) Baseline search space. (b) FAS search space. The meaning of the abbreviations can be referred to 
  Section~\ref{sec:nas}.}
 
\label{fig:Experiment2}
\end{figure}

\subsection{Static vs Static-Dynamic Representation}
\label{sec:strepresentation}

In order to validate whether the dynamic/temporal information are beneficial to spoofing detection, we study the effects of static and dynamic inputs from both separation and fusion views. It can be seen from Fig.~\ref{fig:Experiment1}(b) that only considering dynamic information would lead to performance reduction because of losing much spatial detailed clues. With the addition and normalization fusion strategy, our static-dynamic representation improves 0.3\% ACER compared with static RGB inputs, which proves the effectiveness of temporal context for FAS task. In contrast, directly concatenating static with dynamic causes a slight performance drop.

\noindent\textbf{Static-Dynamic with CDC and CDP.}\quad 
So far, static-dynamic representation is utilized for vanilla networks. It is interesting to explore how central difference family performs under the static-dynamic inputs. As illustrated in Fig.~\ref{fig:Experiment1}(a), with the temporal context, `SD-CDC' and `SD-CDP' are more stable and robust than `CDC' and `CDP', respectively. As rank pooling based dynamic generation can be treated as a special case of spatio-temporal difference, it might be compatible with CDC and CDP based spatial difference. Finally, task-aware positive knowledge (i.e., dynamic-static representation, CDC and CDP) are taken into account into search.


\subsection{NAS with Baseline Search Space}
\label{sec:nasbaseline}

Here, we utilize the latest searching algorithm PC-DARTS~\cite{xu2019pc} and baseline search space as the alternatives. Note that PC-DARTS with similar search space has achieved great performance on generic object classification datasets (e.g., CIFAR-10, CIFAR-100 and ImageNet~\cite{deng2009imagenet}). 

\noindent\textbf{NAS Gap between Object Classification and FAS tasks.}\quad Fig.~\ref{fig:Experiment2}~(a) mainly displays the results of the searched networks with various search space configurations. Firstly, the static ResNet50 (pretrained on ImageNet) with cross-entropy loss (5.2\% ACER) is validated as the non-NAS baseline. Then we search with vanilla convolutions and cross-entropy loss and then try to discover the suitable architecture. However, `NAS van\_CE' fails to detect the spoofing robustly (8.6\% ACER) under slight domain shift case. It indicates that the serious NAS gaps between object classification and FAS task exist. The reasons might be two-folds: 1) the domain shift issues occur between searching/training and testing stage, and 2) the vanilla search space is sub-optimal for FAS task.

\begin{table}[t]

\centering

\caption{The ablation study of the searched architectures on different tasks. Here we follow the same searched networks on CIFAR-10 and ImageNet via PC-DARTS~\cite{xu2019pc}. The results are retrained and tested on Protocol-1 of OULU-NPU.} \label{tab:Necessity}
\resizebox{0.30\textwidth}{!} {\begin{tabular}{l c c c c} 

 \toprule
 Searched on & CDC\&CDP   & ACER(\%)$\downarrow$\\
 \midrule
 CIFAR-10 &   & 9.5\\
CIFAR-10 & $\surd$   & 6.7 \\
  \midrule
 ImageNet  &   & 8.9\\
 ImageNet  & $\surd$ & 5.8 \\
  \midrule
 OULU-NPU &  & 7.6\\
 OULU-NPU  & $\surd$ & 4.5 \\

 \bottomrule
 \end{tabular}}
\end{table}

\noindent\textbf{Impact of Task-Aware Knowledge.}\quad 
Here we explore how domain knowledge (i.e., DeepPixel loss, static-dynamic and central difference family) affects the NAS performance in FAS task. In terms of supervision signals and static-dynamic inputs, both non-NAS `ResNet50-DP' and NAS `NAS Van\_DP' methods benefit from the DeepPixel loss and static-dynamic representation according to Fig.~\ref{fig:Experiment2}~(a). Besides, after introducing CD family into search space, the performance is obviously improved, which even surpasses ResNet50 (`ResNet50\_DP' 4.3\% vs. `NAS CD\_DP' 3.7\% ACER). We use `NAS CD\_DP' with static-dynamic inputs as the default baseline search space setting (named \textbf{NAS-Baseline}) for the following experiments. The searched NAS-Baseline architecture is visualized in Fig.~\ref{fig:searched}(a).

\begin{figure}
\centering
\includegraphics[scale=0.43]{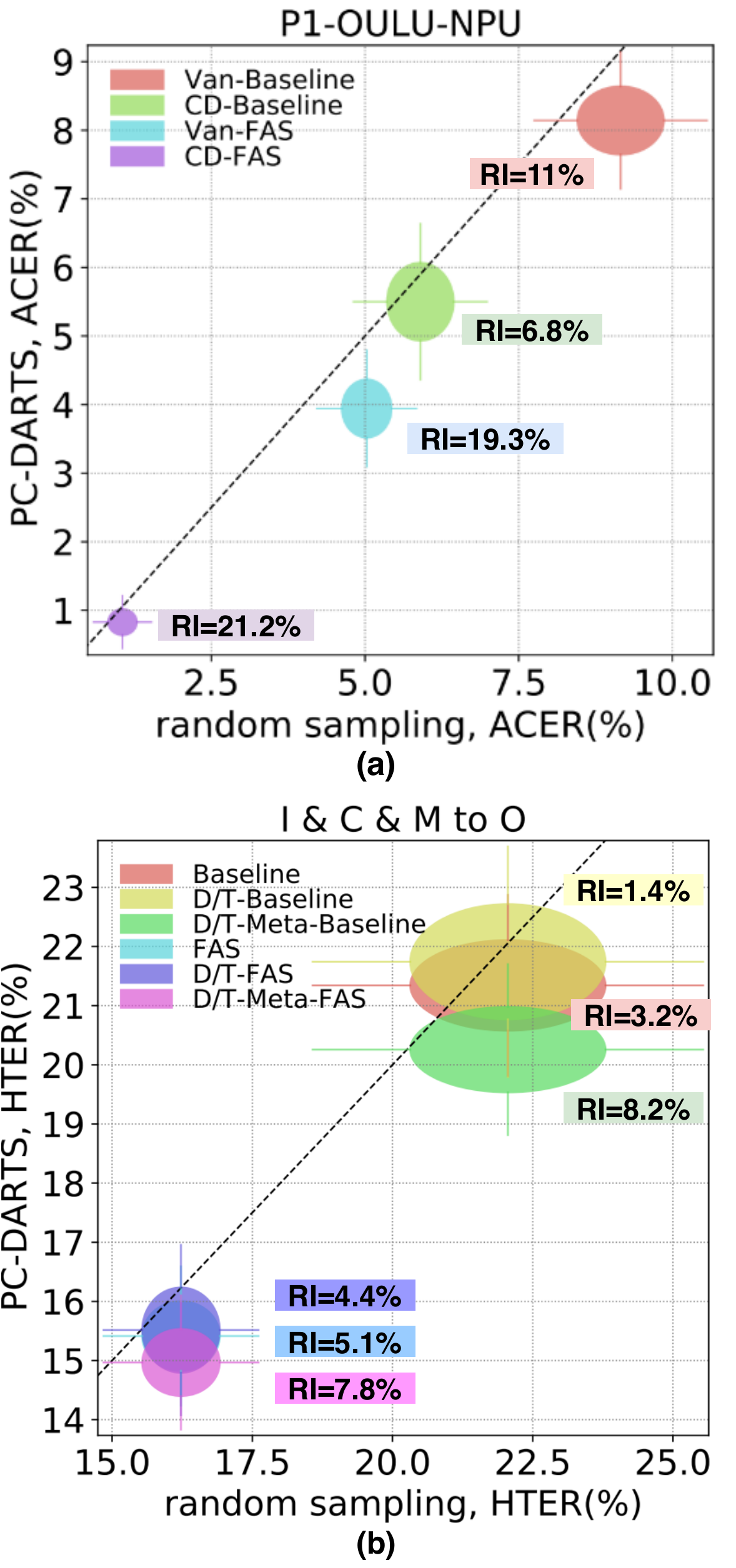}
  \vspace{-0.6em}
  \caption{
  Comparison of PC-DARTS~\cite{xu2019pc} and random sampling from (a) different search spaces on Protocol-1 of OULU-NPU; and (b) cross-dataset testing on OULU-NPU(O) (searching on Replay-Attack(I), CASIA-MFSD(C) and MSU-MFSD(M)). Results lying in the diagonal perform the same as the average architecture, while methods below the diagonal outperform it. 'RI' denotes the relative improvement with respect to random sampling.}
 
\label{fig:RandomSampling}
\end{figure}

\noindent\textbf{Necessity of NAS on FAS Task.}\quad 
It is necessary to investigate how different source tasks (e.g., object classification and FAS tasks) influence the final target task (e.g., FAS task). Table~\ref{tab:Necessity} shows the performance of the searched architectures on CIFAR-10, ImageNet and OULU-NPU (Protocol-1). As the original searched networks on CIFAR-10 and ImageNet via PC-DARTS only consist of vanilla convolution and pooling operators, we also consider to replace them by corresponding CDC or CDP operators. It can be seen from Table~\ref{tab:Necessity} that 1) the architectures searched on object classification tasks are likely to perform worse than those on FAS task (with baseline search space for fair comparison); and 2) CDC\&CDP are helpful for the found architectures on various tasks. Although CDC\&CDP could alleviate such biases, it is still necessary to directly search on FAS task, which is more likely to provide task-aware knowledge for robust searching.

\begin{table}[t]
\centering
\caption{The results of intra testing on OULU-NPU~\cite{Boulkenafet2017OULU}. }
\resizebox{0.50\textwidth}{!}{
\begin{tabular}{|c|c|c|c|c|}

\hline
Prot. & Method & APCER(\%)$\downarrow$ & BPCER(\%)$\downarrow$ & ACER(\%)$\downarrow$ \\
\hline
\multirow{6}{*}{1}
        &GRADIANT ~\cite{boulkenafet2017competition}&1.3 &12.5 & 6.9 \\
        &STASN ~\cite{yang2019face} &1.2 &2.5 & 1.9 \\
        &Auxiliary ~\cite{Liu2018Learning} &1.6 &1.6 & 1.6 \\
        &FaceDs ~\cite{jourabloo2018face} &1.2 &1.7 & 1.5 \\
        &FAS-TD ~\cite{wang2018exploiting} &2.5 &0.0 & 1.3 \\
        &DeepPixBiS ~\cite{george2019deep}&0.8 &0.0 & 0.4 \\
        &CDCN++ ~\cite{yu2020searching}& 0.4 
       & 0.0 & \textbf{0.2} \\
        &\textbf{NAS-Baseline (Ours)} &2.3 &5.1 & 3.7 \\
        &\textbf{NAS-FAS (Ours)} &0.4 &0.0 & \textbf{0.2} \\
\hline
\multirow{6}{*}{2} 
       &DeepPixBiS ~\cite{george2019deep}&11.4 &0.6 & 6.0 \\
       &FaceDs ~\cite{jourabloo2018face}&4.2 &4.4 & 4.3 \\
       &Auxiliary ~\cite{Liu2018Learning}&2.7 &2.7 & 2.7 \\
       &GRADIANT ~\cite{boulkenafet2017competition}&3.1 &1.9 & 2.5 \\
       &STASN ~\cite{yang2019face}&4.2 &0.3 & 2.2 \\
       &FAS-TD ~\cite{wang2018exploiting} &1.7 &2.0 & 1.9 \\
       &CDCN++ ~\cite{yu2020searching}& 1.8
       & 0.8 & 1.3 \\
        &\textbf{NAS-Baseline (Ours)} &3.8 &2.4 & 3.1 \\
        &\textbf{NAS-FAS (Ours)} &1.5
 &0.8 & \textbf{1.2} \\
\hline
\multirow{4}{*}{3} 
       &DeepPixBiS ~\cite{george2019deep}&11.7$\pm$19.6 &10.6$\pm$14.1 & 11.1$\pm$9.4 \\
       &FAS-TD ~\cite{wang2018exploiting}&5.9$\pm$1.9 &5.9$\pm$3.0 & 5.9$\pm$1.0 \\
       &GRADIANT ~\cite{boulkenafet2017competition}&2.6$\pm$3.9 &5.0$\pm$5.3 &3.8$\pm$2.4 \\
       &FaceDs ~\cite{jourabloo2018face}&4.0$\pm$1.8 &3.8$\pm$1.2 &3.6$\pm$1.6 \\
       &Auxiliary ~\cite{Liu2018Learning}&2.7$\pm$1.3 &3.1$\pm$1.7 &{2.9}$\pm$1.5 \\
       &STASN ~\cite{yang2019face}&4.7$\pm$3.9 &0.9$\pm$1.2  &2.8$\pm$1.6 \\
       &CDCN++ ~\cite{yu2020searching}& 1.7$\pm$1.5
       &2.0$\pm$1.2 &1.8$\pm$0.7 \\
       &\textbf{NAS-Baseline (Ours)} &5.2$\pm$2.3 &3.2$\pm$2.0  &4.2$\pm$1.2 \\
        &\textbf{NAS-FAS (Ours)} &2.1$\pm$1.3 &1.4$\pm$1.1  & \textbf{1.7$\pm$0.6} \\
\hline
\multirow{4}{*}{4} 
        &DeepPixBiS ~\cite{george2019deep}&36.7$\pm$29.7 &13.3$\pm$14.1 & 25.0$\pm$12.7 \\
       &GRADIANT ~\cite{boulkenafet2017competition}&5.0$\pm$4.5 &15.0$\pm$7.1 &10.0$\pm$5.0 \\
       &Auxiliary ~\cite{Liu2018Learning}&9.3$\pm$5.6 &10.4$\pm$6.0 &9.5$\pm$6.0 \\
       &FAS-TD ~\cite{wang2018exploiting}&14.2$\pm$8.7 &4.2$\pm$3.8 & 9.2$\pm$3.4 \\
       &STASN ~\cite{yang2019face}&6.7$\pm$10.6 &8.3$\pm$8.4  &7.5$\pm$4.7 \\
       &FaceDs ~\cite{jourabloo2018face}&1.2$\pm$6.3
       &6.1$\pm$5.1 &5.6$\pm$5.7 \\
       &CDCN++ ~\cite{yu2020searching}&4.2$\pm$3.4
       &5.8$\pm$4.9 &5.0$\pm$2.9 \\
       &\textbf{NAS-Baseline (Ours)} &5.2$\pm$2.8 &9.2$\pm$4.6  &8.2$\pm$3.1 \\
       &\textbf{NAS-FAS (Ours)} &4.2$\pm$5.3 &1.7$\pm$2.6  & \textbf{2.9$\pm$2.8} \\
\hline
\end{tabular}
}
\label{tab:OULU}
\end{table}

\subsection{NAS with FAS Search Space}
\label{sec:nasFAS}

As mentioned in Section~\ref{sec:nasbaseline}, the task-aware knowledge is helpful for searching. Therefore, FAS search space is utilized, which consists of stronger task-aware experience (e.g., multi-level features, depth-wise supervision and spatial attention). As illustrated in Fig.~\ref{fig:Experiment2}~(b), we can automatically find the novel architecture `NAS Van\_Max', achieving better performance (0.6\% ACER) than non-NAS `DepthNet' with static-dynamic representation. Furthermore, benefited from refined network space (CDP and spatial attention) and operation space (CDC), we can easily search well-suitable networks to achieve state-of-the-art performance, which can be seen from the columns `NAS CDC\_Max', `NAS CDC\_CDP', `NAS CDC\_Max\_Att' and `NAS CDC\_CDP\_Att'. 

\begin{figure*}
\centering

\includegraphics[scale=0.44]{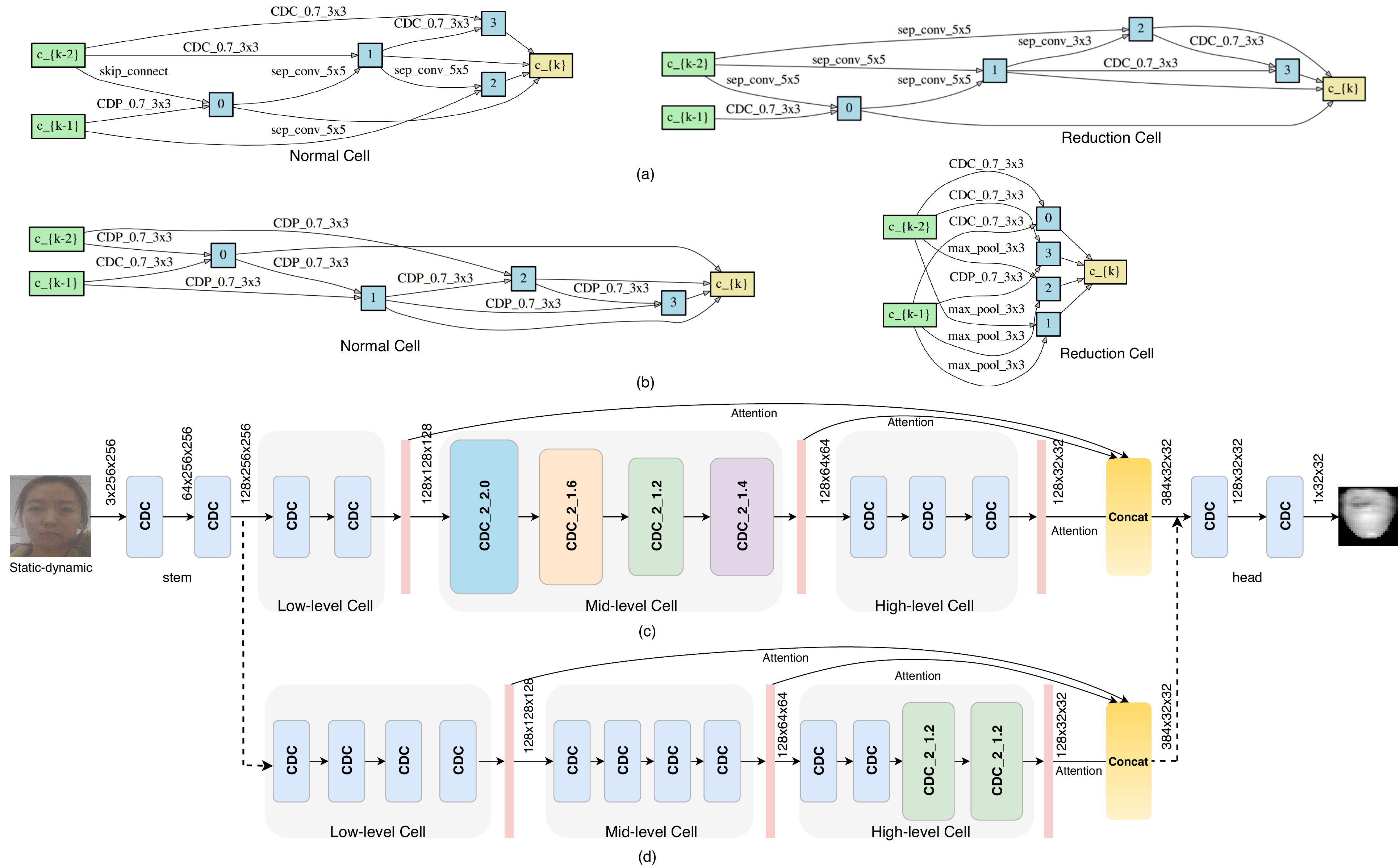}

\vspace{-0.6em}
  \caption{\small{
  Searched neural architectures. (a) Searched network with baseline search space (NAS-Baseline) on Protocol-1 OULU-NPU. (b) NAS-Baseline with Type(Mask Attacks)-aware Meta-NAS on SiW-M. (c) Searched network with FAS search space (NAS-FAS) on Protocol-1 OULU-NPU. Each cell is followed by a CDP layer. (d) NAS-FAS with Type(Mask)-aware Meta-NAS on SiW-M.}
  }
 \vspace{-0.6em}
\label{fig:searched}
\end{figure*}

We use `NAS CDC\_CDP\_Att' with static-dynamic inputs as the default FAS search space setting (named \textbf{NAS-FAS}) for the following experiments. The searched NAS-FAS architecture is shown in Fig.~\ref{fig:searched}~(c). It is interesting to see that the low-level and high-level cells are more compact with shallower and narrower layers while mid-level cell has complex structure with deeper layers.

\subsection{Comparison with Random Sampling}
\label{sec:RandomSampling}

 \noindent\textbf{Under Different Search Spaces.} \quad   To evaluate the efficiency of NAS, we compare it with random sampling in four different search spaces (i.e., 'Van-Baseline', 'CD-Baseline', 'Van-FAS' and 'CD-FAS'). We follow the evaluation metric in~\cite{yang2019evaluation} to calculate a relative improvement over random sampling baseline as $RI = -100\times\left (ACER_{s}-ACER_{r}\right)/ACER_{r}$. RI could offer insights into the quality of the search strategy alone. $ACER_{s}$ and $ACER_{r}$ represent the test performance of the PC-DARTS and random sampling strategies, respectively. Fig.~\ref{fig:RandomSampling}(a) shows the evaluation results on Protocol-1 of OULU-NPU, from which we draw two main conclusions. First, in all 4 search spaces, the PC-DARTS performs consistently better than random sampling ('RI'$>$0), which indicates the simple gradient-based NAS actually helps to discover better-suited architectures. Second, the design of search space influences evaluation a lot. The small range of ACER obtained hints at CD-based search space ('CD-Baseline' and 'CD-FAS'), where even the worst architectures perform reasonably well. This is possibly because CD-based operators enhance the global robustness of the search space.

 \noindent \textbf{Searching on Multiple Domains.} \quad   To evaluate the effectiveness of D/T-Meta-NAS, we compare it with two other settings (NAS and D/T-NAS) on cross-dataset intra-type protocol ('I\&C\&M to \&O' here) with two search spaces (i.e., 'Baseline' and 'FAS'). It can be seen from Fig.~\ref{fig:RandomSampling}(b) that 
1) introducing 'D/T-Meta' improves both Baseline (+5.9\% RI) and FAS (+2.7\% RI) search space dramatically; and 2) without meta-updating of the weights, D/T-Baseline (or D/T-FAS) is even less robust than the original Baseline (or FAS). This is because 'D/T-Meta' exploits the domain shifts knowledge among domain-aware tasks, which provides stable weight initialization for architecture search. Without 'D/T-Meta', the discrepancy from multiple domains/tasks would conflict the optimal search direction.

\subsection{Intra-Dataset Intra-Type Testing}
\label{sec:protocol1}


\noindent\textbf{Results on OULU-NPU.} \quad  As shown in Table~\ref{tab:OULU}, our proposed NAS-FAS ranks first on all 4 protocols (0.2\%, 1.3\%, 1.8\% and 5.0\% ACER, respectively), which indicates the proposed method performs well at the generalization of the external environment, attack mediums and input camera variation. The proposed NAS-FAS outperforms CDCN++~\cite{yu2020searching} by a large margin in the most challenging Protocol-4, which indicates static-dynamic representation and CDP-based search space are beneficial to learn intrinsic spoofing features even with very limited training data. It's worth noting that the searched architecture for NAS-FAS is transferable and generalizes well on all protocols although it is searched on Protocol-1. We also show the results of NAS-Baseline, which achieves acceptable but not SOTA performance, indicating the importance of searching space selection for FAS task.

\begin{table}[t]
\centering
\caption{The results of intra testing on SiW~\cite{Liu2018Learning}. } 
\resizebox{0.49\textwidth}{!}{
\begin{tabular}{|c|c|c|c|c|}
\hline
Prot. & Method & APCER(\%) & BPCER(\%) & ACER(\%) \\
\hline
\multirow{3}{*}{1} 
       &Auxiliary ~\cite{Liu2018Learning}&3.58 &3.58 &3.58 \\
       &STASN ~\cite{yang2019face}&-- &-- &1.00 \\
       &FAS-TD ~\cite{wang2018exploiting} &0.96 &0.50 &0.73 \\
       &CDCN++ ~\cite{yu2020searching}& 0.07
       & 0.17 & \textbf{0.12} \\
        &\textbf{NAS-Baseline (Ours)} &0.34 &1.58 & 0.96 \\
        &\textbf{NAS-FAS (Ours)}&0.07 &0.17 & \textbf{0.12} \\
\hline
\multirow{3}{*}{2} &Auxiliary ~\cite{Liu2018Learning}&0.57$\pm$0.69 &0.57$\pm$0.69 &0.57$\pm$0.69 \\
       &STASN ~\cite{yang2019face}&-- &-- &0.28$\pm$0.05 \\
       &FAS-TD ~\cite{wang2018exploiting}&0.08$\pm$0.14 &0.21$\pm$0.14 & 0.15$\pm$0.14 \\
       &CDCN++ ~\cite{yu2020searching}&0.00$\pm$0.00
       &0.09$\pm$0.10 &\textbf{0.04$\pm$0.05} \\
       &\textbf{NAS-Baseline (Ours)} &0.18$\pm$0.24 &0.28$\pm$0.07  &0.23$\pm$0.18 \\
       &\textbf{NAS-FAS (Ours)} &0.00$\pm$0.00 &0.09$\pm$0.10  & \textbf{0.04$\pm$0.05} \\
\hline
\multirow{3}{*}{3} &STASN ~\cite{yang2019face}&-- &-- &12.10$\pm$1.50 \\
       &Auxiliary ~\cite{Liu2018Learning}&8.31$\pm$3.81 &8.31$\pm$3.80 &8.31$\pm$3.81 \\
       &FAS-TD ~\cite{wang2018exploiting}&3.10$\pm$0.81 &3.09$\pm$0.81 & 3.10$\pm$0.81 \\
      &CDCN++ ~\cite{yu2020searching}&1.97$\pm$0.33
       &1.77$\pm$0.10 &1.90$\pm$0.15 \\
       &\textbf{NAS-Baseline (Ours)} &3.67$\pm$1.04 &7.35$\pm$1.56  &5.51$\pm$1.23 \\
       &\textbf{NAS-FAS (Ours)} &1.58$\pm$0.23 &1.46$\pm$0.08  & \textbf{1.52$\pm$0.13} \\
\hline
\end{tabular}
}
\label{tab:SiW}
\vspace{-1.0em}
\end{table}

\begin{table*}
\centering
\caption{Results of cross-dataset intra-type testing on OULU-NPU, CASIA-MFSD, Replay-Attack, and MSU-MFSD. `w/ D-Meta' denotes searching with Domain-aware Meta-NAS on these four datasets.}

\scalebox{0.96}{\begin{tabular}{c|c|c|c|c|c|c|c|c}
\hline
\multirow{2}{*}{\textbf{Method}} &\multicolumn{2}{c|}{\textbf{O\&C\&I to M}} &\multicolumn{2}{c|}{\textbf{O\&M\&I to C}}&\multicolumn{2}{c|}{\textbf{O\&C\&M to I}} &\multicolumn{2}{c}{\textbf{I\&C\&M to O}} \\
\cline{2-9} &\tabincell{c}{HTER(\%)} &\tabincell{c}{AUC(\%)} &\tabincell{c}{HTER(\%)} &\tabincell{c}{AUC(\%)}&\tabincell{c}{HTER(\%)}&\tabincell{c}{AUC(\%)}&\tabincell{c}{HTER(\%)}&\tabincell{c}{AUC(\%)} \\
\hline
Color Texture ~\cite{boulkenafet2016face} 
& 28.09 & 78.47 & 30.58 & 76.89 & 40.40 & 62.78 & 63.59 & 32.71 \\
\hline
MMD-AAE ~\cite{li2018domain} 
& 27.08 & 83.19 & 44.59 & 58.29 & 31.58 & 75.18 & 40.98 & 63.08 \\
\hline
MADDG ~\cite{shao2019multi} 
& 17.69 & 88.06 & 24.50 & 84.51 & 22.19 & 84.99 & 27.98 & 80.02 \\
\hline
DR-MD-Net~\cite{wang2020cross} 
& 17.02 & 90.10 & 19.68 & 87.43 & 20.87 & 86.72 & 25.02 & 81.47 \\
\hline
RFMeta ~\cite{shao2019regularized} 
& 13.89 & 93.98 & 20.27 & 88.16 & 17.30 & 90.48 & 16.45 & 91.16 \\
\hline
\textbf{NAS-Baseline (Ours)}
& 14.63 & 94.26 & 17.24 & 87.48 & 19.73 & 88.52 & 19.81 & 86.80\\
\hline

\textbf{\tabincell{c}{NAS-Baseline w/ D-Meta (Ours)}}
& \textbf{11.62} & \textbf{95.85} & 16.96 & 89.73 & 16.82 & 91.68 & 18.64 & 88.45\\
\hline

\textbf{NAS-FAS (Ours)}
& 19.53 & 88.63 & 16.54 & 90.18 & 14.51 & 93.84 & 13.80 & 93.43\\
\hline

\textbf{\tabincell{c}{NAS-FAS w/ D-Meta (Ours)}}
& 16.85 & 90.42 & \textbf{15.21} & \textbf{92.64} & \textbf{11.63} & \textbf{96.98} & \textbf{13.16} & \textbf{94.18}\\
\hline

\end{tabular}}

\label{tab:DG}
\end{table*}

\begin{table*}
\centering
\caption{AUC (\%) of the model cross-type testing on CASIA-MFSD, Replay-Attack, and MSU-MFSD.}

\scalebox{0.92}{\begin{tabular}{c|c|c|c|c|c|c|c|c|c|c}
\hline
\multirow{2}{*}{Method} &\multicolumn{3}{c|}{CASIA-MFSD ~\cite{Zhang2012A}} &\multicolumn{3}{c|}{Replay-Attack ~\cite{ReplayAttack}}&\multicolumn{3}{c|}{MSU-MFSD ~\cite{wen2015face}} &\multirow{2}{*}{Overall} \\
\cline{2-10} &\tabincell{c}{Video} &\tabincell{c}{Cut Photo} &\tabincell{c}{Wrapped} &\tabincell{c}{Video}&\tabincell{c}{Digital Photo}&\tabincell{c}{Printed}&\tabincell{c}{Printed}&\tabincell{c}{HR Video}&\tabincell{c}{Mobile Video} & \\
\hline
OC-SVM+BSIF ~\cite{arashloo2017anomaly}
& 70.74 & 60.73 & 95.90 & 84.03 & 88.14 & 73.66 & 64.81 & 87.44 & 74.69 & 78.68$\pm$11.74 \\
\hline
SVM+LBP ~\cite{Boulkenafet2017OULU}
& 91.94 & 91.70 & 84.47 & 99.08 & 98.17 & 87.28 & 47.68 & 99.50
 & 97.61 & 88.55$\pm$16.25 \\
\hline
NN+LBP ~\cite{xiong2018unknown}
& 94.16 & 88.39 & 79.85 & 99.75 & 95.17 & 78.86 & 50.57 & 99.93 & 93.54 & 86.69$\pm$16.25 \\
\hline
DTN ~\cite{liu2019deep}
& 90.0 & 97.3 & 97.5 & 99.9 & 99.9 & 99.6 & \textbf{81.6} & 99.9 & 97.5 & 95.9$\pm$6.2 \\
\hline
\textbf{NAS-Baseline (Ours)}
& 96.32 & 94.86 & 98.6 & 99.46 & 98.34 & 92.78 & 68.31 & 99.89 & 96.76 & 93.9$\pm$9.87 \\
\hline
\textbf{NAS-FAS (Ours)}
&\textbf{99.62} & \textbf{100} & \textbf{100} & \textbf{99.99} & \textbf{99.89} & \textbf{99.98} & 74.62 & \textbf{100.00} & \textbf{99.98} & \textbf{97.12$\pm$8.94} \\
\hline
\end{tabular}
}
\label{tab:cross-type}
\end{table*}

\noindent\textbf{Results on SiW.} \quad   Table~\ref{tab:SiW} compares the performance of our method with four state-of-the-art methods:  Auxiliary~\cite{Liu2018Learning}, STASN~\cite{yang2019face}, FAS-TD~\cite{wang2018exploiting} and CDCN++~\cite{yu2020searching} on SiW dataset. It can be seen from Table~\ref{tab:SiW} that the proposed NAS-FAS performs the best for all three protocols (0.12\%, 0.04\% and 1.52\% ACER, respectively). Specially, with the newly introduced static-dynamic representation and CDP-based search space, NAS-FAS surpasses CDCN++ by 0.38\% ACER on the most challenging Protocol-3 of SiW. The results reveals the excellent generalization capacity of NAS-FAS for 1) face pose and expression; 2) different spoof mediums; and 3) cross presentation attacks.

\subsection{Cross-Dataset Intra-Type Testing}
\label{sec:protocol3}

\begin{figure}
\centering
\includegraphics[scale=0.55]{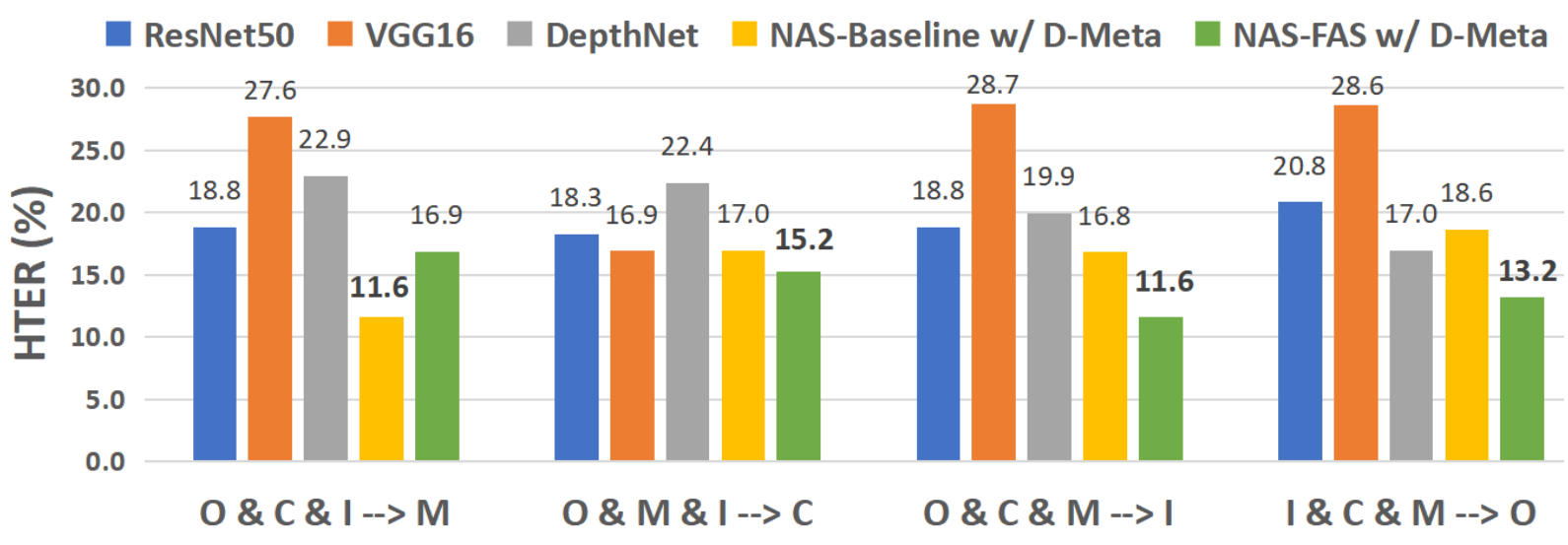}
  \vspace{-1.6em}
  \caption{
  Architecture performance on four cross-dataset intra-type testing sub-protocols. Hand-designed architectures (ResNet50, VGG16 and DepthNet) are likely to perform well when testing on specific few unseen datasets. In contrast, the proposed domain-aware Meta-NAS (yellow and green columns) is helpful for finding generalized and robust architectures for all cross-dataset testings. }
 
\label{fig:DomainMeta}
\end{figure}

Four datasets OULU-NPU (O), CASIA-MFSD (C), Idiap Replay-Attack (I) and MSU-MFSD (M) are utilized here. Specifically, three datasets are randomly selected for training and the remained one leaves for testing. As these four datasets share the same attack types but diverse environments, here we treat each dataset as a specific domain, and search architectures within three known domains with domain(D)-aware Meta-NAS. As shown in Fig.~\ref{fig:DomainMeta}, the fixed human-designed architectures are likely to perform well when testing on specific new unseen datasets. For instance, ResNet50, VGG16 and DepthNet generalize well on MSU-MFSD, CASIA-MFSD and OULU-NPU, respectively. In contrast, the proposed D-Meta-NAS (see yellow and green columns in Fig.~\ref{fig:DomainMeta}) is helpful for discovering robust architectures generalizing on all cross-dataset testings.   

Table \ref{tab:DG} gives the detailed comparisons with the state of the arts. It is clear that 1) despite searching on Protocol-1 of OULU-NPU, the architectures found by NAS-Baseline and NAS-FAS still achieve superior performance under unseen environment; 2) by means of fully exploiting the domain-shifted knowledge, the proposed D-Meta-NAS is able to improve the search quality for both baseline and FAS search space when searching on multiple source domains. Overall, our searched networks in source domains generalize well when testing in unseen target domain. It is surprising that the baseline search space (NAS-Baseline) performs better than the FAS search space (NAS-FAS) in 'O\&C\&I to M' sub-protocol, indicating the biases between task-aware search space and unseen testing domains. In other word, the searched networks only with single task-aware search space are difficult to generalize best for all cases (domains).

\begin{table*}
\centering
\caption{Results of the cross-type testing on SiW-M~\cite{liu2019deep}. `T-Meta' denotes searching with Type-aware Meta-NAS.}

\scalebox{0.75}{\begin{tabular}{c|c|c|c|c|c|c|c|c|c|c|c|c|c|c|c}
\hline
\multirow{2}{*}{Method} &\multirow{2}{*}{Metrics(\%)} &\multirow{2}{*}{Replay} &\multirow{2}{*}{Print} &\multicolumn{5}{c|}{Mask Attacks} &\multicolumn{3}{c|}{Makeup Attacks}&\multicolumn{3}{c|}{Partial Attacks} &\multirow{2}{*}{Average} \\
\cline{5-15} &  &  &  & \tabincell{c}{Half} &\tabincell{c}{Silicone} &\tabincell{c}{Trans.} &\tabincell{c}{Paper}&\tabincell{c}{Manne.}&\tabincell{c}{Obfusc.}&\tabincell{c}{Imperson.}&\tabincell{c}{Cosmetic}&\tabincell{c}{Funny Eye} & \tabincell{c}{Glasses} &\tabincell{c}{Partial} & \\
\hline
\hline

\multirow{4}{*}{SVM$_{RBF}$+LBP~\cite{Boulkenafet2017OULU}} & APCER & 19.1 & 15.4 & 40.8 & 20.3 & 70.3 & 0.0 & 4.6 & 96.9 & 35.3 & 11.3 & 53.3 & 58.5 & 0.6 & 32.8$\pm$29.8 \\
\cline{3-16}  & BPCER & 22.1 & 21.5 & 21.9 & 21.4 & 20.7 & 23.1 & 22.9 & 21.7 & 12.5 & 22.2 & 18.4 & 20.0 & 22.9 & 21.0$\pm$2.9 \\
\cline{3-16}  & ACER & 20.6 & 18.4 & 31.3 & 21.4 & 45.5 & 11.6 & 13.8 & 59.3 & 23.9 & 16.7 & 35.9 & 39.2 & 11.7 & 26.9$\pm$14.5 \\
\cline{3-16}  & EER & 20.8 & 18.6 & 36.3  & 21.4 & 37.2 & 7.5 & 14.1 & 51.2 & 19.8 & 16.1 & 34.4 & 33.0 & 7.9 & 24.5$\pm$12.9 \\

\hline
\hline

\multirow{4}{*}{Auxiliary~\cite{Liu2018Learning}} & APCER & 23.7 & 7.3 & 27.7 & 18.2 & 97.8 & 8.3 & 16.2 & 100.0 & 18.0 & 16.3 & 91.8 & 72.2 & 0.4 & 38.3$\pm$37.4 \\
\cline{3-16}  & BPCER & 10.1 & 6.5 & 10.9 & 11.6 & 6.2 & 7.8 & 9.3 & 11.6 & 9.3 & 7.1 & 6.2 & 8.8 & 10.3 & 8.9$\pm$ 2.0 \\
\cline{3-16}  & ACER & 16.8 & 6.9 & 19.3 & 14.9 & 52.1 & 8.0 & 12.8 & 55.8 & 13.7 & \textbf{11.7} & 49.0 & 40.5 & 5.3 & 23.6$\pm$18.5 \\
\cline{3-16}  & EER & 14.0 & 4.3 & 11.6  & 12.4 & 24.6 & 7.8 & 10.0 & 72.3 & 10.1 & \textbf{9.4} & 21.4 & 18.6 & 4.0 & 17.0$\pm$17.7 \\

\hline
\hline

\multirow{4}{*}{DTN~\cite{liu2019deep}} & APCER & 1.0 & 0.0 & 0.7 & 24.5 & 58.6 & 0.5 & 3.8 & 73.2 & 13.2 & 12.4 & 17.0 & 17.0 & 0.2 & 17.1$\pm$23.3 \\
\cline{3-16}  & BPCER & 18.6 & 11.9 & 29.3 & 12.8 & 13.4 & 8.5 & 23.0 & 11.5 & 9.6 & 16.0 & 21.5 & 22.6 & 16.8 & 16.6 $\pm$6.2 \\
\cline{3-16}  & ACER & 9.8 & \textbf{6.0} & 15.0 & 18.7 & 36.0 & 4.5 & 7.7 & 48.1 & 11.4 & 14.2 & 19.3 & 19.8 & 8.5 & 16.8 $\pm$11.1 \\
\cline{3-16}  & EER & 10.0 & \textbf{2.1} & 14.4 & 18.6 & 26.5 & \textbf{5.7} & 9.6 & 50.2 & 10.1 & 13.2 & 19.8 & 20.5 & 8.8 & 16.1$\pm$ 12.2 \\

\hline
\hline

\multirow{4}{*}{\textbf{NAS-Baseline (Ours)}}& APCER & 23.2 & 14.0 & 12.9 & 20.1 & 26.3 & 15.4 & 10.9 & 50.6 & 12.4 & 13.9 & 36.3 & 37.4 & 9.8 & 21.8 $\pm$12.6 \\
\cline{3-16}  & BPCER  & 16.4 & 8.4 & 9.7 & 12.3 & 22.5 & 5.6 & 8.3 & 33.8 & 5.8 & 11.3 & 27.3 & 23.2 & 3.8 & 14.5$\pm$9.4\\
\cline{3-16}  & ACER & 19.8 & 11.2 & 11.3 & 16.2 & 24.4 & 10.5 & 9.6 & 42.2 & 9.1 & 12.6 & 31.8 & 30.3 & 6.8 & 18.1 $\pm$10.9\\
\cline{3-16}  & EER & 18.6 & 10.8 & 10.9 & 14.9 & 23.3 & 9.6 & 8.4 & 42.8 & 8.2 & 12.4 & 32.4 & 28.2 & 5.1 & 17.4$\pm$ 11.2 \\

\hline
\hline

\multirow{4}{*}{\textbf{\tabincell{c}{NAS-Baseline (Ours) \\ w/ T-Meta}}}& APCER & 10.3 & 14.7 & 20.8 & 17.1 & 17.1 & 5.8 & 7.5 & 31.8 & 0.0 & 16.0 & 22.4 & 24.0 & 5.8 & 14.9 $\pm$8.8 \\
\cline{3-16}  & BPCER  & 11.6 & 10.4 & 18.0 & 14.8 & 8.0 & 4.6 & 9.3 & 30.4 & 1.6 & 17.1 & 20.7 & 23.8 & 6.9 & 13.6$\pm$8.2\\
\cline{3-16}  & ACER & 11.0 & 12.5 & 19.4 & 15.9 & 12.5 & 5.3 & 8.4 & 31.1 & 0.8 & 16.5 & 21.6 & 23.9 & 6.4 & 14.3 $\pm$8.4\\
\cline{3-16}  & EER & 11.3 & 10.4 & 18.6 & 14.8 & 7.9 & 4.8 & 7.5 & 30.4 & 0.0 & 18.0 & 20.7 & 20.6 & 5.8 & 13.1$\pm$ 8.3 \\

\hline
\hline

\multirow{4}{*}{\textbf{NAS-FAS (Ours)}} & APCER & 12.8 & 7.8 & 13.5 & 12.0 & 17.6 & 3.7 & 3.8 & 38.2 & 1.2 & 13.9 & 23.6 & 18.3 & 2.8 & 13$\pm$7.1 \\
\cline{3-16}  & BPCER & 10.4 & 5.4 & 5.7 & 9.0 & 17.0 & 1.5 & 4.8 & 26.4 & 0.4 & 12.9 & 23.2 & 15.9 & 0.8 & 10.3$\pm$8.4\\
\cline{3-16}  & ACER & 11.6 & 6.6 & \textbf{9.6} & \textbf{10.5} & 17.3 & 2.6 & 4.3 & 32.3 & 0.8 & 13.4 & 23.4 & 17.1 & \textbf{1.8} & 11.6$\pm$9.2 \\
\cline{3-16}  & EER & 11.2 & 5.4 & \textbf{6.7}  & \textbf{10.3} & 16.8 & 5.8 & 4.1 & 33.8 & \textbf{0.0} & 14.1 & 23.3 & 15.4 & \textbf{0.6} & 11.3$\pm$9.5 \\

\hline
\hline

\multirow{4}{*}{\textbf{ \tabincell{c}{NAS-FAS (Ours) \\ w/ T-Meta }}} & APCER & 12.8 & 9.0 & 9.7 & 13.1 & 19.1 & 1.1 & 5.4 & 31.0 & 0.0 & 15.0 & 15.1 & 18.6 & 5.0 & 11.9$\pm$8.4 \\
\cline{3-16}  & BPCER & 10.1 & 6.8 & 13.1 & 11.1 & 12.5 & 2.8 & 0.0 & 26.1 & 0.8 & 15.3 & 17.8 & 13.5 & 2.3 & 10.2$\pm$7.5\\
\cline{3-16}  & ACER & \textbf{9.3} & 7.9 & 11.4 & 12.1 & \textbf{15.8} & \textbf{1.9} & \textbf{2.7} & \textbf{28.5} & \textbf{0.4} & 15.1 & \textbf{16.5} & \textbf{16.0} & 3.8 & \textbf{10.9$\pm$7.8} \\
\cline{3-16}  & EER & \textbf{9.3} & 6.8 & 9.7  & 11.1 & \textbf{12.5} & \textbf{2.7} & \textbf{0.0} & \textbf{26.1} & \textbf{0.0} & 15.0 & \textbf{15.1} & \textbf{13.4} & 2.3 & \textbf{9.5$\pm$7.4} \\

\hline
\hline

\end{tabular}
}
\label{tab:SiW-M}
\end{table*}

\subsection{Intra-Dataset Cross-Type Testing}
\label{sec:protocol2}

\noindent\textbf{Results on CASIA-MFSD, Replay-Attack and MSU-MFSD.} \quad  Following the protocols proposed in~\cite{arashloo2017anomaly}, we use CASIA-MFSD, Replay-Attack and MSU-MFSD datasets to perform intra-dataset cross-type testing between replay and print attacks. As shown in Table~\ref{tab:cross-type}, our proposed NAS-FAS achieves the best overall performance (even outperforming the zero-shot learning based method DTN~\cite{liu2019deep}), indicating our searched networks with consistently good generalization ability among unknown attacks. The intrinsic spoofing patterns between seen and unknown attacks might be represented well in our NAS-FAS.

\begin{table*}
\centering
\caption{Results of cross-dataset cross-type testing when trained on OULU-NPU and SiW. The upper and bottom half part denotes the mobile and normal models, respectively. Input size $256\times256\times3$ is utilized for all methods for fair comparisons. `T(Mask)-Meta' denotes searching with Type-aware Meta-NAS on SiW-M without Mask Attacks.}

\scalebox{0.90}{\begin{tabular}{c|c|c|c|c|c|c|c|c}
\hline
\multirow{2}{*}{Method} &\multirow{2}{*}{\#Params} 
&\multirow{2}{*}{\#FLOPs}&\multicolumn{2}{c|}{\textbf{3DMAD~\cite{erdogmus2014spoofing}}} &\multicolumn{2}{c|}{\textbf{HKBU-MARs~\cite{liu20163d}}}&\multicolumn{2}{c}{\textbf{CASIA-SURF 3DMask (Ous)}}  \\
\cline{4-9} &  &  &  \tabincell{c}{AUC(\%)}&\tabincell{c}{HTER(\%)}  &\tabincell{c}{AUC(\%)}&\tabincell{c}{HTER(\%)}&\tabincell{c}{AUC(\%)}&\tabincell{c}{HTER(\%)}\\

\hline
MobileNetV1 ~\cite{howard2017mobilenets} & 3.20 M & 1.48 G
& 98.64 & 8.51  & 94.35 & 15.63  & 54.00 & 46.00 \\

MobileNetV2 ~\cite{sandler2018mobilenetv2} & 2.21 M & 816.27 M
& 95.58 & 9.85  & 75.75 & 33.37  & 68.85 & 39.71 \\

MobileNetV3 ~\cite{howard2019searching} & 4.21 M & 582.95 M
& \textbf{99.68} & 0.29  & 79.78 & 28.35  & 66.69 & 40.10 \\

ShuffleNetV2 ~\cite{ma2018shufflenet} & 2.27 M & 394.18 M
& 98.75 & 5.76  & 67.61 & 35.94  & 55.33 & 47.50 \\

PNAS ~\cite{liu2018progressive} & 3.63 M & 345.53 M
& 92.27 & 16.73  & 77.86 & 22.05  & 73.29 & 37.95 \\

\tabincell{c}{NAS S-Van-CE (PC-DARTS~\cite{xu2019pc})}  & 0.71 M & 140.49 M
& 97.72 & 2.82  & 85.43 & 25.12 & 54.64 & 46.13 \\
\textbf{NAS-Baseline (Ours)} & 2.57 M & 398.72 M
& 99.31 & \textbf{0.22}  & 88.91 & 15.13  & 72.83 & 37.68 \\
\textbf{\tabincell{c}{NAS-Baseline w/ T(Mask)-Meta (Ours)}} & 1.27 M & 212.34 M
& 99.46 & 0.48  & 92.52 & 12.94  & 75.76 & 32.22 \\

\hline

ResNet50 ~\cite{he2016deep} & 23.52 M & 4.08 G
& 99.06 & 1.47  & 87.15 & 22.66  & 52.16 & 48.34 \\
DepthNet ~\cite{Liu2018Learning}  & 2.22 M & 93.14 G
& 99.04 & 0.29  & 88.32 & 14.64  & 60.44 & 32.54 \\
DTN ~\cite{liu2019deep} & 1.33 M & 26.41 G
& 98.86 & 1.47 & 91.01 & 6.47 & 69.24 & 38.97 \\


\textbf{NAS-FAS (Ours)} & 2.94 M & 52.67 G
& 99.18 & 0.26  & 93.21 & \textbf{5.86}  & 83.91 & 16.46 \\

\textbf{\tabincell{c}{NAS-FAS  w/ T(Mask)-Meta (Ours)}} & 2.58 M & 53.96 G
& 99.08 & 1.18  & \textbf{94.84} & 6.75  & \textbf{85.78} & \textbf{15.00} \\
\hline
\end{tabular}}

\label{tab:cross-cross}
\end{table*}

\noindent\textbf{Results on SiW-M.} \quad  Following the same cross-type testing protocol (13 attacks leave-one-out) on SiW-M, we compare our proposed methods with three recent FAS methods~\cite{Boulkenafet2017OULU,Liu2018Learning,liu2019deep} to validate the generalization capacity of unseen attacks. Besides searching directly on Protocol-1 OULU-NPU, we also consider searching type-aware architectures on SiW-M due to its rich attack types. To be specific, five macro type definitions (i.e., Replay, Print, Mask, Makeup and Partial) with leave-one-out setting are utilized for Type(T)-aware Meta-NAS. For instance, for the sub-protocol of target 'Silicone' type, we meta search architectures on 4 macro source types (Replay, Print, Makeup and Partial).  

As shown in Table ~\ref{tab:SiW-M}, our NAS-FAS achieves an overall better ACER and EER, with the improvement over the previous state-of-the-art~\cite{liu2019deep} by 24\% and 26\% respectively. Specifically, we detect almost all 'Impersonation' and 'Partial Paper' attacks (EER$<$1\%) while the previous methods perform poorly on 'Impersonation' attack. Furthermore, equipped with 'T-Meta', the searched architectures generalize better on both Baseline (4.3\% EER reduced) and FAS search spaces (1.8\% EER decreased). Although 'NAS-FAS w/ T-Meta' achieves best overall performance, it still performs worse than 'NAS-Baseline' in a few sub-protocols (e.g., 'Print' and 'Half'). This is possibly because of the biased attention and conflicts when meta-searching on multiple type-shifted tasks. Thus, one possible future direction is to design more robust NAS for unseen attack type detection.

The type-aware meta-searched architectures (on Replay, Print, Makeup and Partial attacks, and without Mask attacks) with Baseline and FAS search space are visualized in Fig.~\ref{fig:searched}(b) and (d), respectively. We name these two architectures as 'NAS-Baseline w/ T(Mask)-Meta' and 'NAS-FAS w/ T(Mask)-Meta', respectively, which are also used for the following cross-dataset cross-type testing. Compared with the architectures searched on Protocol-1 OULU-NPU (Fig.~\ref{fig:searched}(a)(c)), 'NAS-Baseline w/ T(Mask)-Meta' has more CD-based operators while 'NAS-FAS w/ T(Mask)-Meta' has heavier high-level cell.


\subsection{Cross-Dataset Cross-Type Testing}
\label{sec:cross-cross}
In this new proposed protocol, large-scale data from OULU-NPU and SiW training sets are used for training and then 3DMAD, HKBU-MARs and the proposed 3DMask are utilized for testing. It is challenging as there are only two most common presentation attack types (i.e., print and replay) in training set but testing on unseen domain with unseen mask attacks. Table \ref{tab:cross-cross} shows that 3DMAD is the easiest and all methods could achieve excellent performance (above 92\% AUC). It is worth noting that NAS-Baseline performs better than NAS-FAS in 3DMAD, where the faces are with low resolution and high compression. In other words, the Baseline search space is likely to be more practical in unknown severe environment.  Moreover, all methods also generalize well (more than 85\% AUC) on HKBU-MARs. However, limited quality of the 3DMAD and HKBU-MARs datasets (laboratory controlled environment and unrealistic mask appearance) makes evaluation more difficult because nearly all methods hold similarly good performance. 

In contrast, our proposed 3DMask dataset is more challenging and similar to complex real-world indoor and outdoor scenarios. As a result, some classical models like ResNet~\cite{he2016deep}, DepthNet~\cite{Liu2018Learning} and DTN~\cite{liu2019deep} could only obtain less than 70\% AUC, which is still far from the requirement of practical use. Although our NAS-FAS could achieve less than 20\% HTER in 3DMask, it is also far from being desired. We hope 3DMask and this new protocol could benefit FAS community for future research. 

At last, we analyze the impact of unseen domain and attack for NAS. It can be seen from Table \ref{tab:cross-cross} that searching by PC-DARTS~\cite{xu2019pc} with Baseline search space, `NAS S-Van-CE' performs near worst among all methods on three mask datasets, which also indicates weak transfer capacity of the searched network when searching on very different domain and attack types. Our proposed NAS-Baseline and NAS-FAS devote to solve this issue via introducing task-aware search space and achieve significantly better performance. Furthermore, slight improvement could be obtained when introducing Type(Mask)-aware Meta-NAS, which validates the effectiveness of our search method when searching on multiple source domains and types and evaluating on unseen target ones.

\begin{figure}
\centering
\includegraphics[scale=0.45, trim=38 0 0 0, clip]{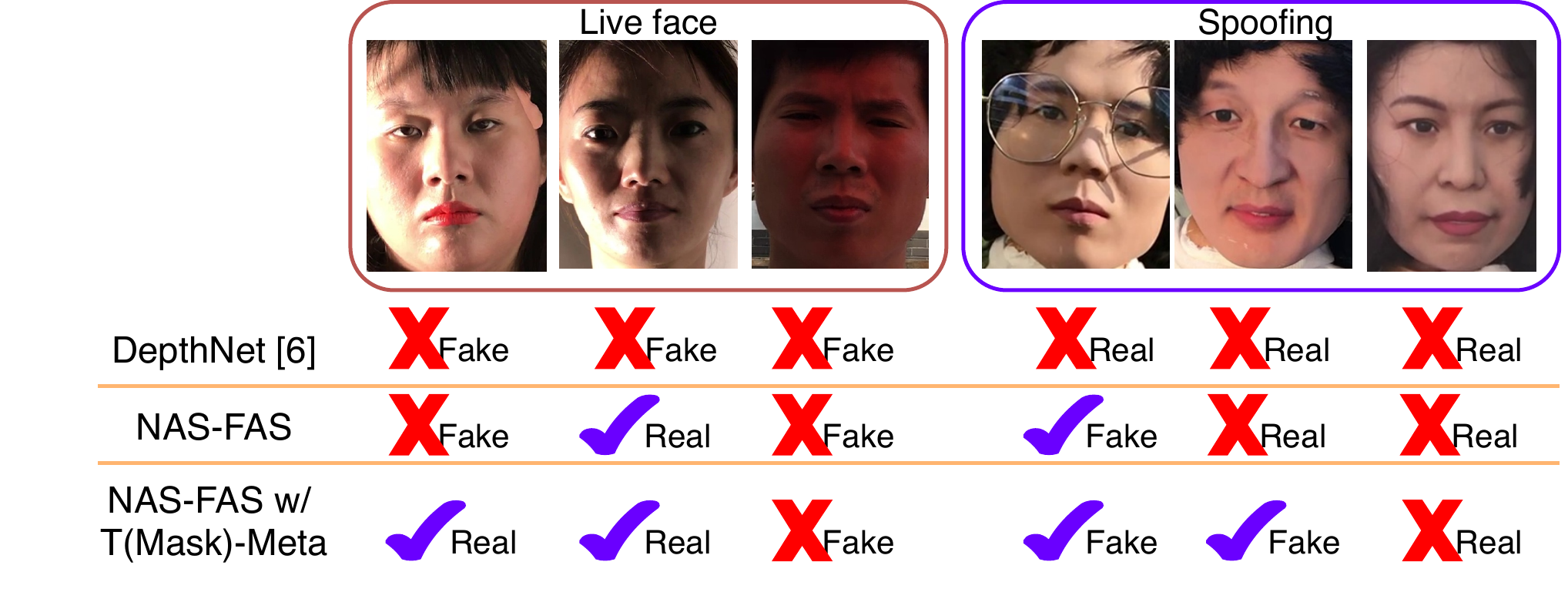}
  \vspace{-1.0em}
  \caption{
  Qualitative analysis on 3DMask. The items 'Real' and 'Fake' are the predicted results from the models.}
 
\label{fig:Visualization}
\vspace{-0.5em}
\end{figure}

\noindent\textbf{Model Size and Computational Cost.} \quad   Table \ref{tab:cross-cross} displays the model size and computational cost of both normal and mobile settings. Compared with the well-known mobile models (e.g., MobileNetV1, MobileNetV2, MobileNetV3, ShuffleNetV2 and PNAS), the proposed 'NAS-Baseline  w/ T(Mask)-Meta' trades-off the efficiency and precision better. It has only 1.27M parameters and 212.34M FLOPs but achieves great performance on all three unseen mask datasets. In terms of the models with normal setting, our 'NAS-FAS  w/ T(Mask)-Meta' has similar parameters and FLOPs with the famous 'DepthNet~\cite{Liu2018Learning}' but outperforms the state-of-the-art methods for a large margin. In the future, resource constraints could be considered for searching more lightweight and robust architectures.

\noindent\textbf{Qualitative Analysis.} \quad  Here we visualize a few hard samples on the CASIA-SURF 3DMask dataset. It can be seen from Fig. \ref{fig:Visualization} that the 'DepthNet~\cite{Liu2018Learning}' is easily to encounter false-reject and false-accept results when the scenarios for the live faces are challenging and the spoofing masks are with high fidelity, respectively. In contrast, our proposed methods are more robust to the environment illumination changes, and more likely to detect the 3D masks with manufacture artifacts or smooth texture.

\section{Conclusion} \label{sec:conclusion}
In this paper, we propose the first neural architecture search (NAS) for face anti-spoofing (FAS) task. Task-aware knowledge is applied for search space design, including static-dynamic representation, central difference convolution and pooling operations. Moreover, domain/type-aware Meta-NAS is proposed for discovering generalized and robust architectures on multiple source datasets and attack types. In order to validate the transferability of NAS for FAS task, we establish a `cross-dataset cross-type' testing protocol with new dataset `CASIA-SURF 3DMask'. Extensive experiments are conducted to verify the effectiveness of our methods. 


As this work mainly focuses on studying the impacts of search space for FAS task, one of the main contributions is to find the excellent task-aware search space. Thus, based on our proposed search space, future directions include: 1) searching optimal $\theta$ and $\lambda$ for CD-based operators in diverse layers/channels; 2) searching multi-branch (e.g., static and dynamic branches, multi-modality) networks for FAS task.



\noindent\textbf{Acknowledgments} \quad This work was supported by the Academy of Finland for project MiGA (grant 316765), ICT 2023 project (grant 328115), Infotech Oulu, and the Chinese National Natural Science Foundation Projects $\#$61961160704, $\#$61876179, Science and Technology Development Fund of Macau No. 0025/2019/A1. The authors also wish to acknowledge CSC-IT Center for Science, Finland.



\ifCLASSOPTIONcaptionsoff
  \newpage
\fi



%
\bibliographystyle{IEEEtran}
\bibliography{IEEEabrv,reference}

%

\begin{IEEEbiography}[{\includegraphics[width=1in,height=1.25in,clip,keepaspectratio]{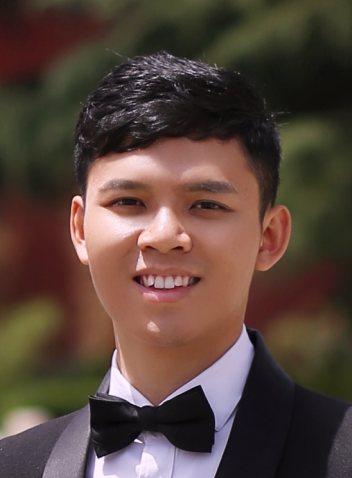}}]{Zitong Yu}
 received the M.S. degree from University of Nantes, France, in 2016, and he is currently a Ph.D. candidate in the Center for Machine Vision and Signal Analysis, University of Oulu, Finland. His research interests focus on remote photoplethysmograph measurement, face anti-spoofing and video understanding.
\end{IEEEbiography}

\begin{IEEEbiography}[{\includegraphics[width=1in,height=1.25in,clip,keepaspectratio]{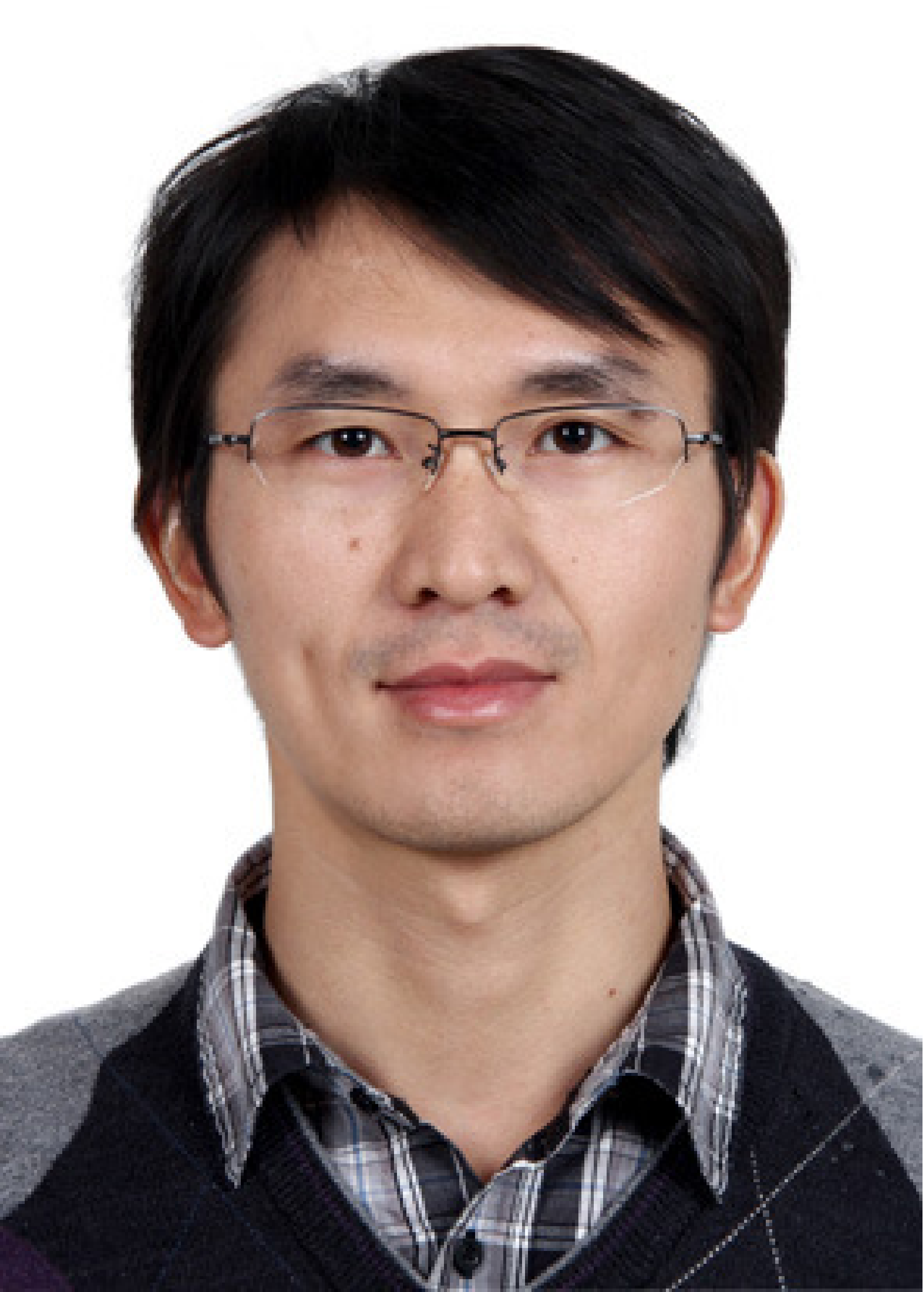}}]{Jun Wan}
received the BS degree from the China University of Geosciences, Beijing, China, in 2008, and the PhD degree from the Institute of Information Science, Beijing Jiaotong University, Beijing, China, in 2015. Since January 2015, he has been a Faculty Member with the National Laboratory of Pattern Recognition, Institute of Automation, Chinese Academy of Science, China, where he currently serves as an Associate Professor. 
He is an associate editor of the IET Biometrics. He has served as co-editor of special issues in TPAMI, MVP and Entropy.
\end{IEEEbiography}

\begin{IEEEbiography}[{\includegraphics[width=1in,height=1.25in,clip,keepaspectratio]{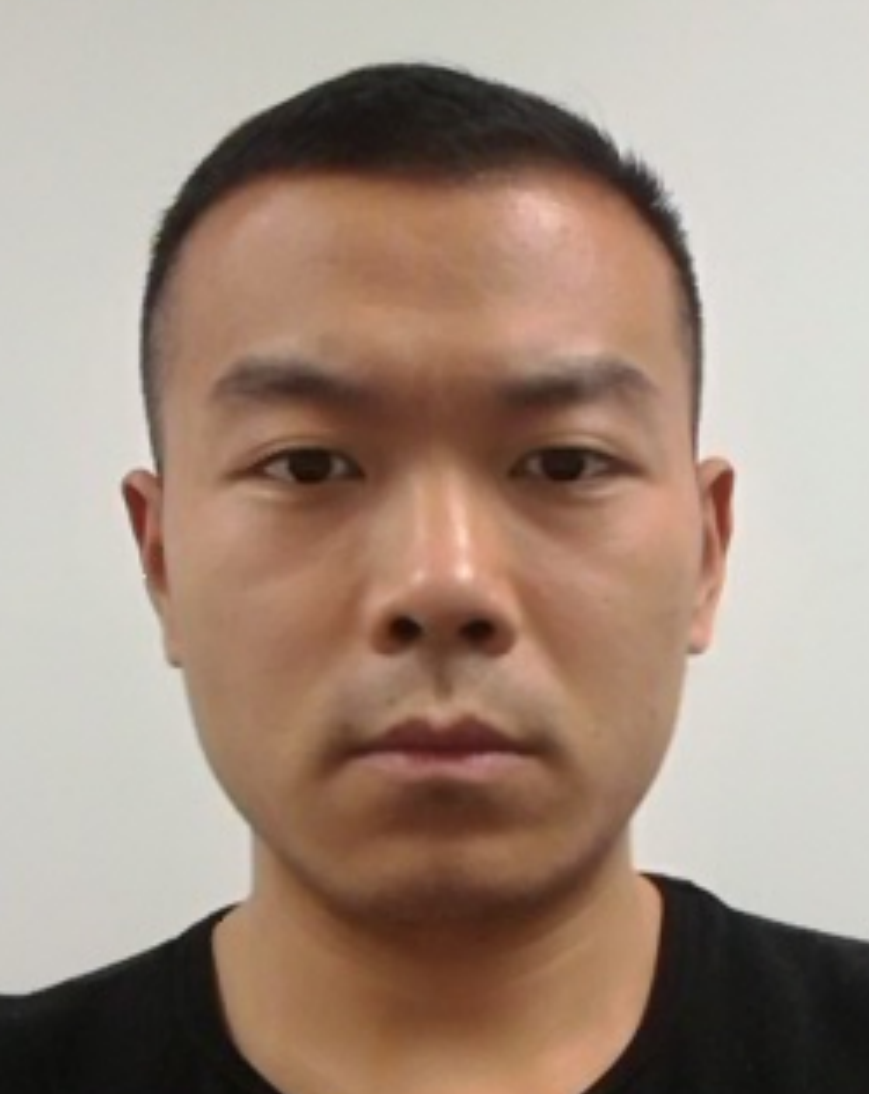}}]{Yunxiao Qin}
 received the M.S. degree in Control Science and Engineering from Northwestern Polytechnical University, Xian, China, in 2015, where he is currently pursuing the Ph.D. degree. His current research interests include computer vision, meta-learning and deep reinforcement learning.

\end{IEEEbiography}

\begin{IEEEbiography}[{\includegraphics[width=1in,height=1.25in,clip,keepaspectratio]{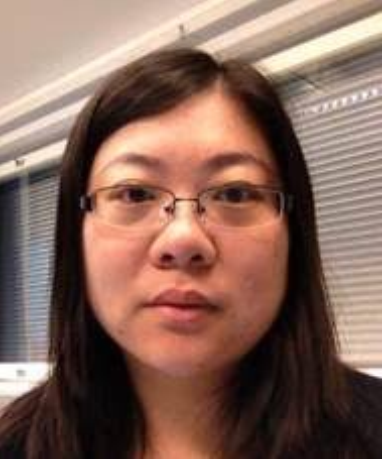}}]{Xiaobai Li}
 received her B.Sc degree in Psychology from Peking University, M.Sc degree in Biophysics from the Chinese Academy of Science, and Ph.D. degree in Computer Science from University of Oulu. She is currently an assistant professor in the Center for Machine Vision and Signal Analysis of University of Oulu. Her research of interests include spontaneous vs. posed facial expression comparison, micro-expression and deceitful behaviors, and heart rate measurement from facial videos. 

\end{IEEEbiography}

\begin{IEEEbiography}[{\includegraphics[width=1in,height=1.25in,clip,keepaspectratio]{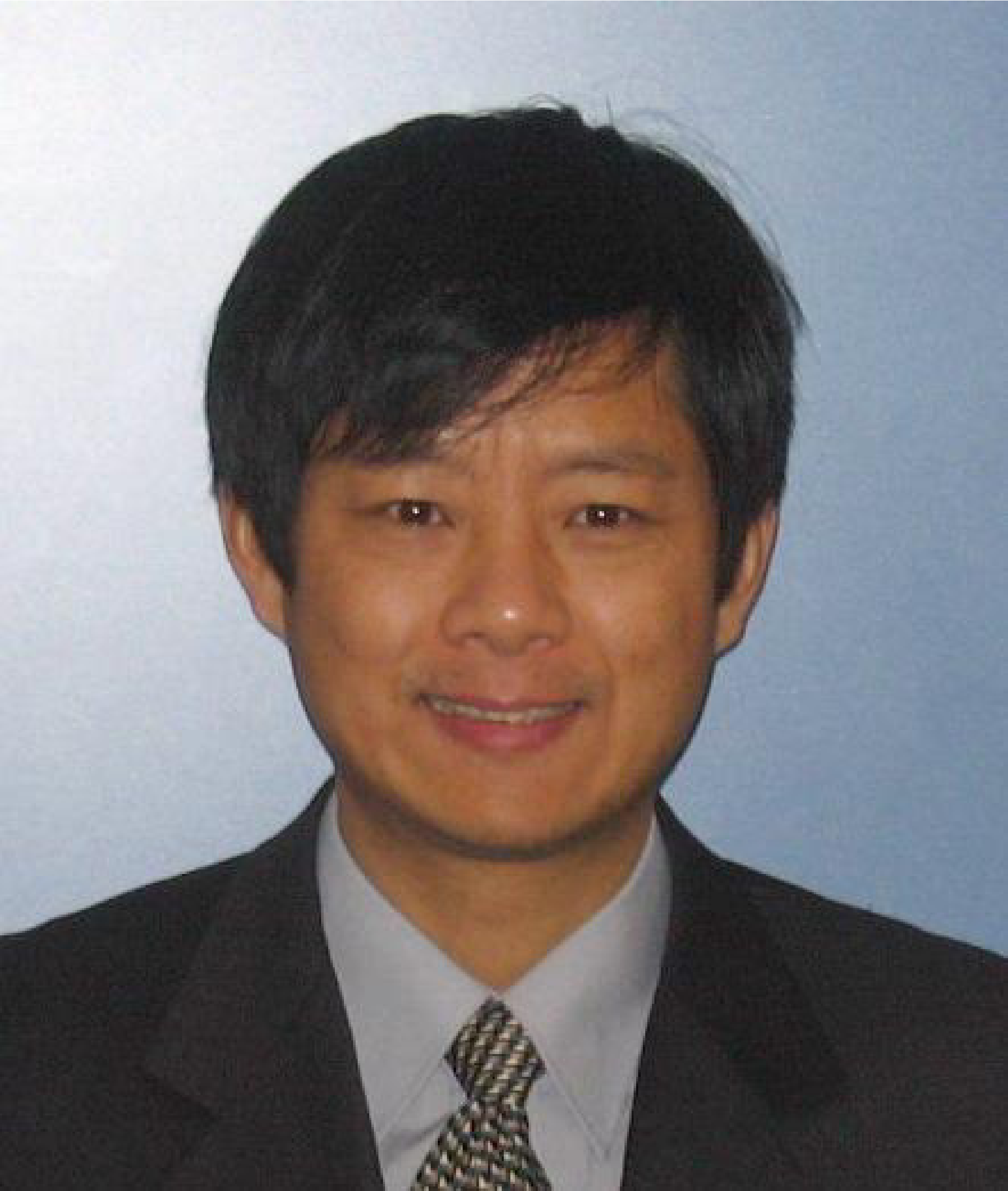}}]{Stan Z. Li}
 (Fellow, IEEE) received the B.Eng. degree from Hunan University, China, in 1982, the M.Eng. degree from the National University of Defense Technology, China, in 1985, and the Ph.D. degree from Surrey University, U.K., in 1991. He is a Chair Professor of Artificial Intelligence with Westlake University, China. He was a Researcher and the Director of the Center for Biometrics and Security Research with Institute of Automation, Chinese Academy of Sciences. He was a Researcher with Microsoft Research Asia and an Associate Professor with Nanyang Technological University, Singapore. He has published over 500 papers with Google scholar index of over 42000 and h-index of 95.
\end{IEEEbiography}
\begin{IEEEbiography}[{\includegraphics[width=1in,height=1.25in,clip,keepaspectratio]{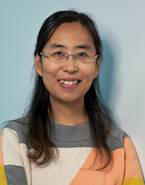}}]{Guoying Zhao}
  (SM'12) is currently a Professor with the Center for Machine Vision and Signal Analysis, University of Oulu, Finland, where she has been a senior researcher since 2005 and an Associate Professor since 2014. She received the Ph.D. degree in computer science from the Chinese Academy of Sciences, Beijing, China, in 2005. She has authored or co-authored more than 240 papers in journals and conferences. Her papers have currently over 13580 citations in Google Scholar (h-index 53). She is co-program chair for ACM International Conference on Multimodal Interaction (ICMI 2021), was co-publicity chair for FG 2018, General chair of 3rd ICBEA 2019, and Late Breaking Results Co-Chairs of 21st ACM ICMI 2019, has served as area chairs for several conferences and is associate editor for Pattern Recognition, IEEE TCSVT, and Image and Vision Computing Journals. She has lectured tutorials at ICPR 2006, ICCV 2009, SCIA 2013 and FG 2018, authored/edited three books and eight special issues in journals. Dr. Zhao was a Co-Chair of many International Workshops at ICCV, CVPR, ECCV, ACCV and BMVC. 
\end{IEEEbiography}




\end{document}